\documentclass{article}

\usepackage{microtype}
\usepackage{graphicx}
\usepackage{subcaption}
\usepackage{booktabs}
\usepackage{multirow}
\usepackage{makecell}

\usepackage{hyperref}
\usepackage{capt-of}

\usepackage[preprint]{icml2026}
\usepackage[utf8]{inputenc}
\usepackage[T1]{fontenc}

\usepackage{amsmath}
\usepackage{amssymb}
\usepackage{mathtools}
\usepackage{amsthm}

\usepackage[capitalize,noabbrev]{cleveref}

\theoremstyle{plain}

\theoremstyle{definition}

\theoremstyle{remark}

\usepackage[textsize=tiny]{todonotes}

\icmltitlerunning{XSPLAIN: XAI-enabling Splat-based Prototype Learning for Attribute-aware INterpretability}

\begin{document}
\def\our{XSPLAIN}
\def\fullmname{XAI-enabling Splat-based Prototype Learning for Attribute-aware INterpretability}

%

\twocolumn[
  \icmltitle{XSPLAIN: XAI-enabling Splat-based Prototype Learning \\ for Attribute-aware INterpretability}



  \icmlsetsymbol{equal}{*}

  \begin{icmlauthorlist}
    \icmlauthor{Dominik Galus}{wr}
    \icmlauthor{Julia Farganus}{wr}
    \icmlauthor{Tymoteusz Zapala}{wr}
    \icmlauthor{Miko{\l}aj Czachorowski}{wr}
    \icmlauthor{Piotr Borycki}{yyy}
    \icmlauthor{Przemys{\l}aw Spurek}{yyy,sch}
    \icmlauthor{Piotr Syga}{wr}
  \end{icmlauthorlist}

  \icmlaffiliation{wr}{Wrocław University of Science and Technology}
  \icmlaffiliation{yyy}{Jagiellonian University}
  \icmlaffiliation{sch}{IDEAS Research Institute}

  \icmlcorrespondingauthor{D. Galus}{279690@student.pwr.edu.pl}

  \icmlkeywords{Machine Learning, ICML}

    \vskip 0.3in

]

\begin{figure*}
\includegraphics[width=0.99\linewidth]{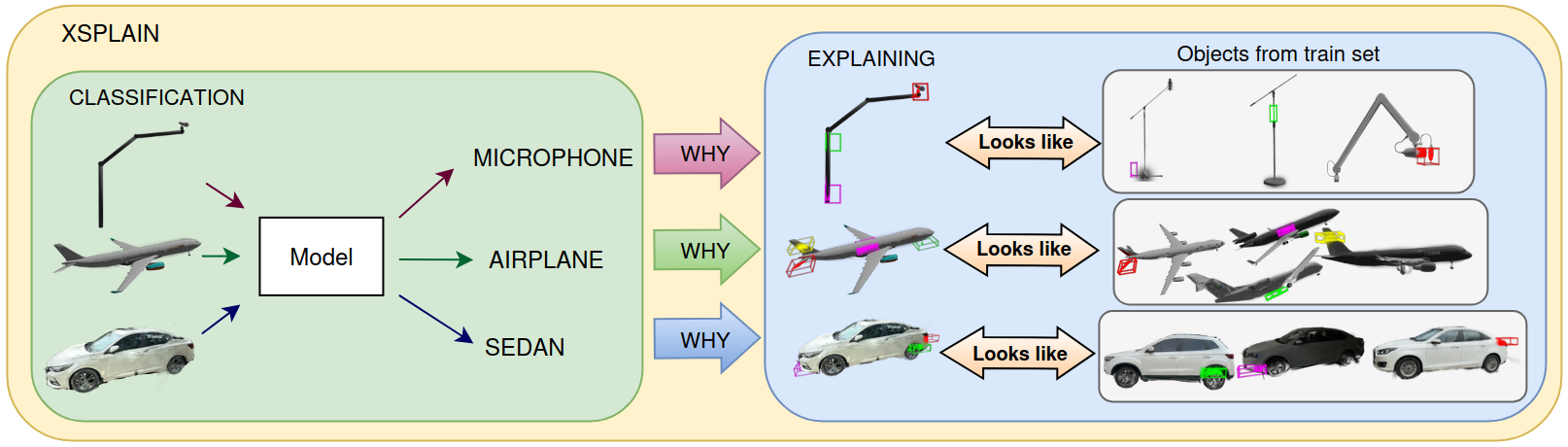}
    \caption{ \textbf{XSPLAIN} provides ante-hoc, prototype-based explanations for 3D Gaussian Splat classification. A PointNet-based classifier predicts the object category from Gaussian Splat representations, while identifying the most influential voxel regions that drive the decision. Explanations are generated by retrieving representative training examples that activate similar latent responses in the same regions, enabling intuitive  ``looks like that'' reasoning grounded in both geometry and semantic attributes.}
  \label{fig:teaser}
\end{figure*}




\printAffiliationsAndNotice{}  


\begin{abstract} 3D Gaussian Splatting (3DGS) has rapidly become a standard for high-fidelity 3D reconstruction, yet its adoption in multiple critical domains is hindered by the lack of interpretability of the generation models as well as classification of the Splats. While explainability methods exist for other 3D representations, like point clouds, they typically rely on ambiguous saliency maps that fail to capture the volumetric coherence of Gaussian primitives. We introduce \our{}, the first ante-hoc, prototype-based interpretability framework designed specifically for 3DGS classification. Our approach leverages a voxel-aggregated PointNet backbone and a novel, invertible orthogonal transformation that disentangles feature channels for interpretability while strictly preserving the original decision boundaries. Explanations are grounded in representative training examples, enabling intuitive ``this looks like that'' reasoning without any degradation in classification performance. A rigorous user study (N=51) demonstrates a decisive preference for our approach: participants selected \our{} explanations 48.4\% of the time as the best, significantly outperforming baselines $(p<0.001)$, showing that \our{} provides transparency and user trust.
\footnote{The source code for this work is available at: 
\href{https://github.com/Solvro/ml-splat-xai}{GitHub}.}
\end{abstract}



\section{Introduction}

Recent years have observed an accretion of deep learning methods dedicated to 3D due to growing data acquisition, mainly in fields such as robotic applications~\citep{keetha2024splatam}, autonomous driving~\citep{zhou2024drivinggaussian}, augmented reality and medical imaging~\citep{li2024endo}. Although they achieve impressive performance, their decision-making processes remain largely unclear or rely on spurious correlations \cite{buhrmester2019analysisexplainersblackbox}. To overcome these limitations, a class of solutions has been proposed under the umbrella of eXplainable AI (XAI)~\cite{xu2019explainable} to provide transparency and interpretability. Generally, these algorithms are grouped into two categories: post-hoc approaches and ante-hoc approaches, where the model is designed to be interpretable by nature. Post-hoc methods serve pre-trained models without altering their architecture, such as SHAP \cite{lundberg2017unified}, LIME \cite{ribeiro2016should}, LRP \cite{bach2015pixel}, and Grad-CAM \cite{selvaraju2020grad} 
However, they usually focus on a single instance and provide saliency maps that may not provide decisive insights. 
Ante-hoc models incorporate interpretable architectures and usually use \textit{prototypes} \cite{chen2019looks,nauta2023pip,struski2024infodisent}, which are beneficial for explaining a given phenomenon by relating it to the dataset used to train the model. Although significant progress has been made in explaining 2D image classifiers, explainability in 3D remains considerably less explored, despite the increasing adoption of 3D representations in real-world systems.

Recently, 3D Gaussian Splatting (3DGS) \cite{kerbl20233dgaussiansplattingrealtime} has emerged as a powerful representation for modeling complex geometry, offering an efficient and continuous alternative to point clouds and meshes \cite{10870258,chen2025survey3dgaussiansplatting}. Although 3DGS has been extensively studied in the context of rendering and reconstruction, its use for discriminative tasks such as classification \cite{zhang2025mitigating} and, more importantly, the interpretability of such models, has received limited attention. Existing explanation methods for 3D data often rely on point-level saliency or gradient-based attribution, which can be noisy, difficult to interpret, and do not leverage the spatial coherence inherent in Gaussian-based representations.

In this work, we address these limitations by introducing \textbf{\our{}}, an ante-hoc, prototype-based explainable framework for the classification of objects represented by 3D Gaussian primitives. Our \our{} can operate as a standalone classifier, as the interpretability mechanism is a plug-in module that does not influence classification performance. Our method produces explanations that are both \emph{faithful} to the model’s decision process and \emph{interpretable} to humans by establishing the predictions in spatially localized and semantically consistent regions of the 3D representation.

At a high level, \our{} builds on a PointNet-inspired \cite{qi2017pointnetdeeplearningpoint} backbone with a voxel aggregation module that preserves spatial structure while remaining permutation-invariant. The model is trained in two stages: first, the backbone is optimized purely for classification,  second, the trained backbone is frozen, and a learnable, invertible transformation is optimized to disentangle feature channels for interpretability while strictly preserving the original decision boundaries. Building on this disentangled representation, \our{} employs a prototype-based interpretability mechanism. For each feature channel, representative training examples are identified and used as prototypes, allowing explanations to be formed by direct comparison between regions of the test sample and analogous regions in the training data. Explanations are thus expressed as localized subsets of Gaussian primitives, offering intuitive, example-based insights into the model’s predictions.

We evaluate \our{} on multiple 3D Gaussian Splatting classification benchmarks, focusing on subsets of the ShapeSplat \citep{ma2024shapesplat} dataset and the MVImageNet-GS \citep{zhang2025mitigating} benchmark. Our experiments demonstrate that \our{} maintains competitive classification performance while producing coherent, spatially grounded explanations. Qualitative results show that the identified regions correspond to meaningful object parts, while quantitative analyzes validate the disentanglement and stability of the learned representations.

In summary, our contributions are threefold:
\begin{itemize}
  \item We introduce the first ante-hoc, prototype-based explainability method for classification models operating on 3D Gaussian Splatting representations.
  \item We propose a PointNet-inspired architecture and a stage-wise feature disentanglement strategy guided by a purity objective, enabling spatially coherent, semantically isolated explanations.
  \item We conduct a comprehensive evaluation on multiple 3D Gaussian Splatting datasets, comparing \our{} against existing post-hoc explainability methods and demonstrating improved interpretability while maintaining competitive classification performance.
\end{itemize}

\section{Related Work}

\textbf{Prototype-based explainability} In safety-critical applications, explanations are required to verify that decisions are grounded in meaningful characteristics rather than spurious cues. While some methods examine models post-hoc \citep{crabbe2021explaining}, intrinsic (ante-hoc) interpretability \citep{koh2020concept} aims to build transparent models from scratch. Prototype learning for deep neural networks usually involves finding structures in latent space representations. ProtoPNet \citep{chen2019looks} introduces learnable prototypes for each class in the spirit of 'this looks like that' reasoning, which agrees with human perception, where the final prediction is a weighted sum of prototype scores. As prototypes usually exhibit similarity, ProtoPShare \citep{10.1145/3447548.3467245} introduces a merge-pruning mechanism to merge redundant prototypes, improving stability and prediction accuracy. 
Other extensions focus on utilizing classifiers other than linear ones, like the decision tree in ProtoTree \citep{Nauta2020NeuralPT} or KNN in ProtoKNN \citep{ukai2023protoknn}. The applicability of prototypes extends beyond images, covering motion analysis (ProtoFormer; \citealp{pmlr-v235-han24d}) or tabular data (ProtoGate; \citealp{10.5555/3692070.3692948}).
Deformable ProtoPNet \citep{donnelly2022deformable} and TesNet \citep{9709918} introduce orthogonality constraints to further enhance prototypes flexibility and transparency. InfoDisent \citep{struski2024infodisent} is a hybrid approach utilizing a trainable orthogonal transformation on pixel space to enable disentanglement of feature space, thus associating each channel with a prototype. EPIC \citep{borycki2025epicexplanationpretrainedimage} introduces post-hoc feature channel disentanglement to find representative patches for the most active channels in the sample.

\textbf{Point Cloud Interpretability} Intrinsic methods introduce specific architecture design, like XPCC \citep{9714048} with prototype based classifier that utilizes kernel point CNN to extract features and class-weighted similarity to representative prototypes for the final decision, Interpretable3D \citep{Feng_Quan_Wang_Wang_Yang_2024} that optimizes prototypes based on the cosine similarity to the correct predictions or C-PointNet \citep{Zhang2019Explaining} which modifies the former architecture by replacing the max pooling layer with a class-attentive feature module, assigning each point to the class that maximizes the activation value. InfoCons \citep{li2025infocons} applies information-theoretic approach to extract a group of interpretable critical concepts. On the other hand, post-hoc methods are usually model-agnostic, like LIME3D \citep{9706626} that explains a point cloud by fitting a surrogate model on a perturbed original sample, Feature Based Interpretability introduced in \citep{levi2024fastsimpleexplainabilitypoint} that evaluates per-point contribution based on pre-bottleneck features' probing, or BubblEX \citep{9846966} that introduces interpretability module based on GradCAM to generate saliency maps conditioned on a class. PointMask \citep{DBLP:journals/corr/abs-2007-04525} incorporates a differentiable masking layer to mask out points with negligible contribution. They also show that PointNet architecture is prone to learn from bias pattern in training dataset. 

\textbf{Gaussian Splatting Interpretability} 3DisGS \citep{zhang2025interpretable} proposes a generative framework for single-view reconstruction with feature disentanglement at both coarse and fine-grained levels, such that each dimension encodes an interpretable semantic factor. However, it is an unsupervised approach with an image input, without clear relevance to the classification on pure 3D objects presented in this work. 

\begin{figure*}[t]
  \centering
  \includegraphics[width=\textwidth]{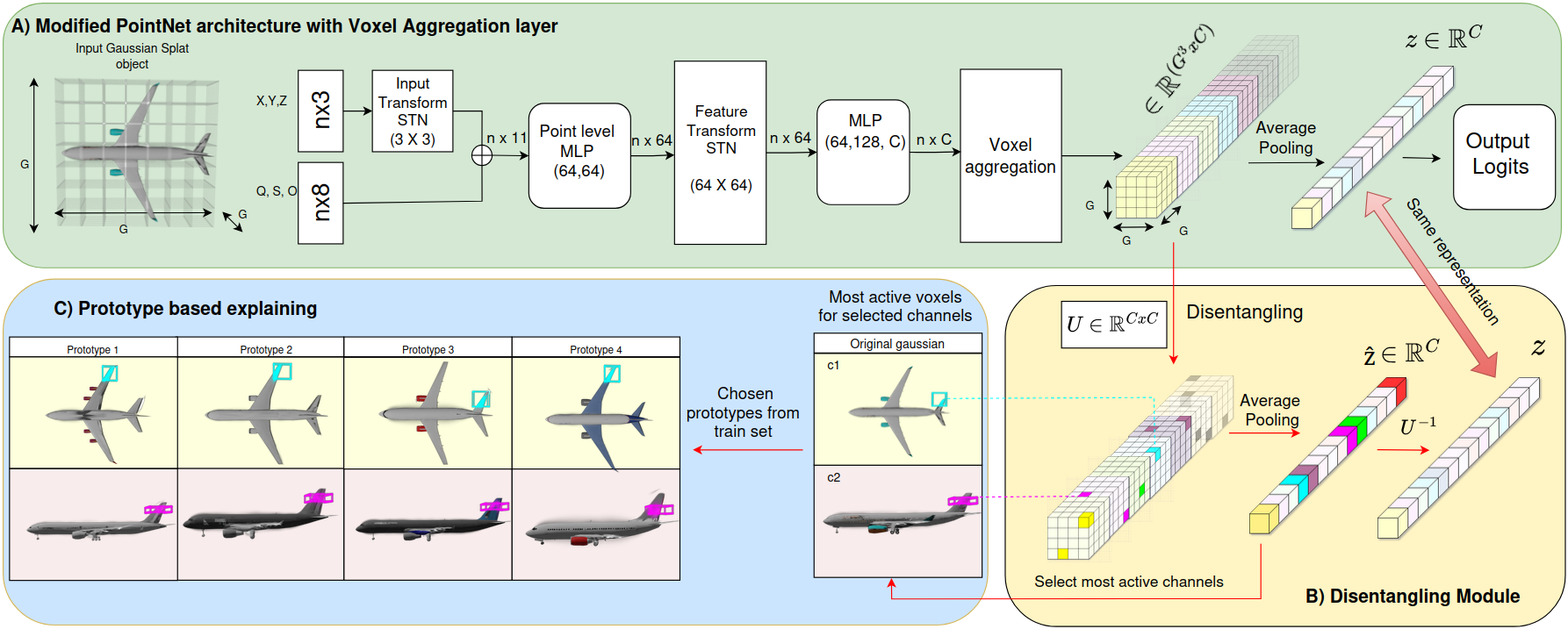}
  \caption{\textbf{Overview of the XSPLAIN architecture}
A) The classification backbone is a modified PointNet architecture extended by a voxel aggregation layer, producing structured latent representations at the voxel level from Gaussian Splat inputs. B) An attachable disentangling module learns an invertible linear transformation that separates latent channels for interpretability while preserving the original global representation and classification output. C) Explanations are generated by identifyingthe most active disentangled channels, visualizing the corresponding influential voxels, and retrieving representative training examples that exhibit similar channel activations. 
}
  \label{fig:xsplain-arch}
  \vspace{-0.3cm} 
\end{figure*}

\section{XSPLAIN}
Our framework consists of 3 components: (1) a backbone network inspired by PointNet~\citep{qi2017pointnetdeeplearningpoint} with a voxel aggregation module, (2) a trainable parametrized matrix for feature disentanglement, and (3) a prototype-based interpretability mechanism. Crucially, our approach follows a two-stage training procedure: the backbone is trained for classification, then frozen, while the parametrized matrix is optimized for interpretability. The architecture and the explaining procedure are illustrated in~Fig.~\ref{fig:xsplain-arch}.

\paragraph{Problem Formulation}
Let $\mathcal{G} = \{g_i\}_{i=1}^{N}$ denote a 3D Gaussian Splatting representation with $N$ Gaussian primitives. We consider only geometric attributes, excluding view-dependent color:

\begin{equation}
g_i = (\mathbf{x}_i, \mathbf{s}_i, \mathbf{q}_i, \alpha_i) \in (\mathbb{R}^{3}, \mathbb{R}^{3}, \mathbb{R}^{4}, \mathbb{R}^{1})~,
\end{equation}
where \(\mathbf{x}_i\) denotes the 3D position, \(\mathbf{s}_i\) the anisotropic scaling factors, \(\mathbf{q}_i\) the orientation quaternion, and \(\alpha_i\) the opacity. This
choice is motivated by recent findings~\cite{zhang2025mitigating} showing that geometric attributes alone encode rich structural semantics, while excluding color improves
robustness to appearance variations and reduces input dimensionality. This formulation allows our method to generalize to scenarios where color information may be unavailable or unreliable, e.g., objects reconstructed under varying illumination or from depth sensors.


\paragraph{Backbone architecture}
Our backbone is based on the PointNet architecture~\citep{qi2017pointnetdeeplearningpoint}.
We employ two Spatial Transformer Networks for input alignment: one operating in 3D coordinate space and another in the 64-dimensional feature space-followed by shared 1D convolutions that produce per-point feature vectors $\mathbf{f}_i \in \mathbb{R}^{C}$. The key modification we introduce is a \textbf{voxel aggregation module} that replaces the global max-pooling operation, providing spatial structure for interpretability while improving robustness to outliers.

\noindent \textbf{\textit{Voxel Aggregation Module}}
To establish a spatially structured representation suitable for interpretability, we employ an \textbf{intrinsic partitioning strategy}. Specifically, we partition the normalized input space into a regular grid of $G^3$ voxels. Crucially, the voxel index $v(i)$ for each Gaussian primitive is assigned \textit{prior} to the Spatial Transformer Network based on its initial coordinates:
\begin{equation}
v(i) = \lfloor \tilde{x}_i \cdot G \rfloor \cdot G^2 + \lfloor \tilde{y}_i \cdot G \rfloor \cdot G + \lfloor \tilde{z}_i \cdot G \rfloor~,
\end{equation}
where coordinates are clipped to $[0, G-1]$.
This design choice ensures that explanations are anchored in the object's original coordinate frame rather than a learned canonical pose. As a result, the spatial attribution directly reflects where the model attends in the input space, providing transparent insight into positional dependencies of the classifier.

This index $v(i)$ serves as immutable metadata bound to each primitive throughout the network. Point features are subsequently aggregated within each voxel using max-pooling:
\begin{equation}
\mathbf{h}_v = \max_{i: v(i)=v} \mathbf{f}_i \in \mathbb{R}^{C}.
\end{equation}

\textbf{\textit{Structure Preservation via Decoupling}} This architectural choice decouples local feature aggregation from global semantic integration. Unlike standard PointNet~\citep{qi2017pointnetdeeplearningpoint} which collapses all spatial information via global max-pooling, our voxel strategy preserves the association between features and their coarse spatial locations.

\noindent \textbf{\textit{Global Feature and Classification}}
We obtain the global feature vector by applying average pooling across all voxels:
\begin{equation}
  \mathbf{z} = \frac{1}{G^3} \sum_{v=1}^{G^3} \mathbf{h}_v \in \mathbb{R}^{C}.
\end{equation}
The final classification is performed by a linear layer:
\begin{equation}
  \hat{y} = \arg\max_k \left(\mathbf{W}_{\text{cls}} \mathbf{z}\right)_k,
\end{equation}
where $\mathbf{W}_{\text{cls}} \in \mathbb{R}^{K \times C}$ are the classifier weights.

\paragraph{Two-Stage Training Procedure}
Although the architecture is intrinsically interpretable (ante-hoc), training is performed in two stages to decouple classification from interpretability. Joint optimization of accuracy and prototype purity can reduce discriminative power. Our decoupled strategy preserves backbone feature learning and predictive accuracy while ensuring output-level faithfulness in the second stage.

\noindent \textbf{\textit{Stage 1: Backbone Training}} The entire backbone network is trained $f_\theta$ (including both STNs, convolutional layers, voxel aggregation, and the classification head) end-to-end using a combination of classification loss and a density-aware regularization term.

\noindent \textbf{\textit{Classification Loss}}
The primary objective is the standard cross-entropy loss:
\begin{equation}
  \mathcal{L}_{\text{cls}} = -\frac{1}{|\mathcal{D}|} \sum_{(\mathcal{G}, y) \in \mathcal{D}} \log \frac{\exp((\mathbf{W}_{\text{cls}}
\mathbf{z})_y)}{\sum_{k=1}^{K} \exp((\mathbf{W}_{\text{cls}} \mathbf{z})_k)},
\end{equation}
where $\mathcal{D}$ denotes the training dataset.

\noindent \textbf{\textit{Density-Aware Regularization}}
The limitation of PointNet-style architectures with max-pooling aggregation is their tendency to focus on isolated outlier points rather than dense, geometrically meaningful regions. In our voxel-based setting, this manifests as high activations in sparsely populated voxels containing only a few Gaussian primitives, which often correspond to reconstruction artifacts or noise rather than salient object parts.

To address this, we introduce a regularization term based on the Kullback-Leibler divergence that encourages the network to align its attention with the spatial density of the input. Let $\mathbf{a} \in \mathbb{R}^{G^3}$ denote the voxel activation magnitudes computed as $a_v = \|\mathbf{h}_v\|_2$, and let
$\mathbf{n} \in \mathbb{R}^{G^3}$ denote the point counts per voxel, where $n_v = |\{i : v(i) = v\}|$.

We define two probability distributions over voxels. The activation distribution is obtained by applying a temperature-scaled softmax:
\begin{equation}
  p_v = \frac{\exp(\text{ReLU}(a_v) / \tau)}{\sum_{u=1}^{G^3} \exp(\text{ReLU}(a_u) / \tau)},
\end{equation}
where $\tau > 0$ is a temperature hyperparameter controlling the sharpness of the distribution. The target density distribution is derived from the point counts:
\begin{equation}
  q_v = \frac{n_v^\beta + \epsilon}{\sum_{u=1}^{G^3} (n_u^\beta + \epsilon)},
\end{equation}
where $\epsilon=1 \cdot e^{-6}$ ensures numerical stability and $\beta > 0$ controls sensitivity to density differences

The density-aware regularization is defined as the KL divergence between the target and activation distributions:
\begin{equation}
  \mathcal{L}_{\text{density}} = D_{\text{KL}}(p \| q) = \sum_{v=1}^{G^3} p_v \log \frac{p_v}{q_v}.
\end{equation}

Minimizing this term encourages the model to assign higher activations to voxels with more Gaussian primitives. Intuitively, densely populated voxels are more likely to represent coherent geometric structures (e.g., object surfaces, distinctive parts) rather than isolated outliers or reconstruction noise. This regularization guides the network toward learning features grounded in the actual geometry of the object.

\noindent \textbf{\textit{Total Training Objective}}
The complete training objective for Stage 1 combines both terms:
\begin{equation}
  \mathcal{L}_{\text{Stage 1}} = \mathcal{L}_{\text{cls}} + \lambda \mathcal{L}_{\text{density}}~,
\end{equation}
where $\lambda \geq 0$ is a hyperparameter balancing classification accuracy and density alignment. This formulation ensures that the learned voxel representations are both discriminative for classification and focused on geometrically meaningful regions, which subsequently benefits the interpretability of the prototype-based explanations in Stage 2.

\noindent \textbf{\textit{Stage 2: Parametrized Matrix Training}}
In the second stage, the backbone parameters are frozen and fixed to the solution obtained in the first stage:
\begin{equation}
  \theta^* = \arg\min_\theta \mathcal{L}_{\text{cls}}(\theta)~, 
\end{equation}
and no updates of \(\theta\) are performed during training.

We then introduce a trainable parametrized matrix $\mathbf{U} \in \mathbb{R}^{C \times C}$  inserted between the voxel aggregation module and the global pooling operation. Only the parameters of this matrix are optimized during the second stage on a dynamically updated set of prototype examples (detailed in \ref{prototypes_selection}). In this phase, the backbone remains fixed. This design choice ensures that: (1)~the classification performance established in Stage 1 is preserved, and (2)~the interpretability module does not interfere with learned feature representations.

\paragraph{Parametrized Feature Disentanglement}
We introduce a transformation matrix $\mathbf{U} \in \mathbb{R}^{C \times C}$ that maps voxel features into a disentangled representation where channels capture semantically distinct information:
\begin{equation}
  \tilde{\mathbf{H}} = \mathbf{U} \mathbf{H}_{\text{flat}}~,
\end{equation}
where $\mathbf{H}_{\text{flat}} \in \mathbb{R}^{C \times G^3}$ is the flattened voxel tensor.

To ensure that the transformation preserves the classifier outputs by construction, we restrict it to an orthogonal mapping in the feature space, which preserves inner products and Euclidean distances prior to the fixed classification layer.

Let $\mathbf{P} \in \mathbb{R}^{C \times C}$ be a learnable, unconstrained weight matrix. We derive a skew-symmetric matrix $\mathbf{A}$ and the disentanglement matrix $\mathbf{U}$ as:
\begin{align}
    \mathbf{A} &= \mathbf{P} - \mathbf{P}^T. \\
    \mathbf{U} &= \exp(\mathbf{A}).
\end{align}

Since $\mathbf{A}$ is skew-symmetric ($\mathbf{A}^T = -\mathbf{A}$), the resulting matrix $\mathbf{U}$ is guaranteed to be orthogonal ($\mathbf{U}^T \mathbf{U} = \mathbf{I}$). Moreover, $\det(\mathbf{U}) = 1$, ensuring the transformation is volume-preserving and strictly invertible. Initializing $\mathbf{P} = \mathbf{0}$, yields $\mathbf{U} = \mathbf{I}$, ensuring the second training stage begins with the exact feature space learned by the backbone.

\noindent \textbf{\textit{Classifier Compensation}}
After optimizing $\mathbf{P}$ for interpretability, we compensate the classification head to strictly preserve the original model predictions. Since $\mathbf{U}$ is orthogonal,
its inverse is simply its transpose ($\mathbf{U}^{-1} = \mathbf{U}^T$). We compute the adjusted weights as:
\begin{equation}
\mathbf{W}'_{\text{cls}} = \mathbf{W}_{\text{cls}} \mathbf{U}^{T}.
\end{equation}
This formulation allows us to rotate the internal feature representation for human interpretation without altering the decision boundary (see
Appendix~\ref{app:classifier_compensation} for proof).

\paragraph{Prototype-Based Interpretability} \label{prototypes_selection}
Our interpretability mechanism identifies prototype examples that maximally activate each feature channel.

\noindent \textbf{\textit{Prototype Discovery}}
For each channel $c \in \{1, \ldots, C\}$, we identify the top-$k$ training samples with the highest activation. For a sample with transformed voxel features
$\tilde{\mathbf{H}}$, we compute:
\begin{equation}
  a_c = \max_{v: n_v > 0} \tilde{h}_{c,v},
\end{equation}
where $\tilde{h}_{c,v}$ is the activation of the $c$ -th channel in voxel $v$, and $n_v$ is the count of points. We maintain prototype indices $\mathcal{P}_c = \{j_1,
\ldots, j_k\}$ for the samples with highest $a_c$.

\noindent \textbf{\textit{Purity Metric}}
To train $\mathbf{U}$ toward disentangled representations, we introduce a purity metric measuring how well each channel is isolated in its maximally activated
voxel. For a sample associated with channel $c$:
\begin{equation}
  v^* = \arg\max_{v: n_v > 0} \tilde{h}_{c,v}, \qquad
  \text{purity}_c = \frac{\tilde{h}_{c,v^*}}{\|\tilde{\mathbf{h}}_{v^*}\|_2 + \epsilon}.
\end{equation}
A purity close to 1 indicates that the channel $c$ dominates the representation in its most activated voxel. The training objective maximizes average purity:
\begin{equation}
  \mathcal{L}_{\text{purity}} = -\frac{1}{|\mathcal{B}|} \sum_{(i, c) \in \mathcal{B}} \text{purity}_c^{(i)},
\end{equation}
where $\mathcal{B}$ is a batch of prototype-channel pairs.

\noindent \textbf{\textit{Dynamic Prototype Updates}}
We periodically recompute prototypes during training, using a curriculum that reduces the number per channel:
\begin{equation}
  k_t = \left\lfloor k_{\text{init}} - \frac{t}{T}(k_{\text{init}} - k_{\text{final}}) \right\rfloor,
\end{equation}
for epoch $t$, allowing the matrix to first capture broad patterns before focusing on representative examples.

\paragraph{Explanation Generation}
Given a test sample $\mathcal{G}^*$, we identify the influential channels for the predicted class.

\noindent \textbf{\textit{Channel Importance}}
The importance of channel $c$ for the predicted class $\hat{y}$ is:
\begin{equation}
  \text{importance}_c = w'_{\hat{y},c} \cdot \text{ReLU}(\tilde{z}^*_c),
\end{equation}
where $w'_{\hat{y},c}$ is the compensated classifier weight. We select the top-$m$ channels.

\noindent \textbf{\textit{Spatial Localization}}
For each important channel $c$, we localize the explanation by finding the maximally activated voxel index:
\begin{equation}
  v^*_c = \arg\max_{v} \tilde{h}^*_{c,v}.
\end{equation}
A key advantage of our intrinsic partitioning strategy is the direct reversibility of the spatial attribution. Since the voxel assignment $v(i)$ for each primitive is determined at the input stage and remains invariant to the STN's alignment transformations, we do not require inverse geometric projections to locate the relevant features. The explanatory subset of primitives $\mathcal{E}_c$ is obtained via a direct lookup:
\begin{equation}
  \mathcal{E}_c = \{ g_i \in \mathcal{G}^* \mid v(i) = v^*_c \}.
\end{equation}
We visualize the explanation by rendering only the subset $\mathcal{E}_c$, allowing for an exact and artifact-free presentation of the regions driving the model's decision.

\begin{table}[t]
\centering
\begin{tabular}{l l c}
\hline
\textbf{Dataset} & \textbf{Model} & \textbf{Accuracy} \\
\hline
Toys        & PointNet   & 0.865 \\
Toys        & PointNet++ & 0.934 \\
Toys        & PointNeXt  & 0.898 \\
Toys        & PointMLP   & 0.870 \\
Toys        & PointNet + Vox Agg   & 0.899 \\
\hline
MACGS       & PointNet   & 0.873 \\
MACGS       & PointNet++ & 0.871 \\
MACGS       & PointNeXt  & 0.805 \\
MACGS       & PointMLP   & 0.898 \\
MACGS       & PointNet + Vox Agg  & 0.818 \\
\hline
Shapesplat & PointNet   & 0.869 \\
Shapesplat & PointNet++ & 0.875 \\
Shapesplat & PointNeXt  & 0.875 \\
Shapesplat & PointMLP   & 0.803 \\
Shapesplat & PointNet + Vox Agg   & 0.880 \\
\hline
\end{tabular}
\caption{Baseline results using supervised point cloud models with additional GS inputs}
\label{tab:gs_pointcloud_benchmark}
\end{table}

\begin{figure}[t]
  \centering
  \includegraphics[width=\columnwidth]{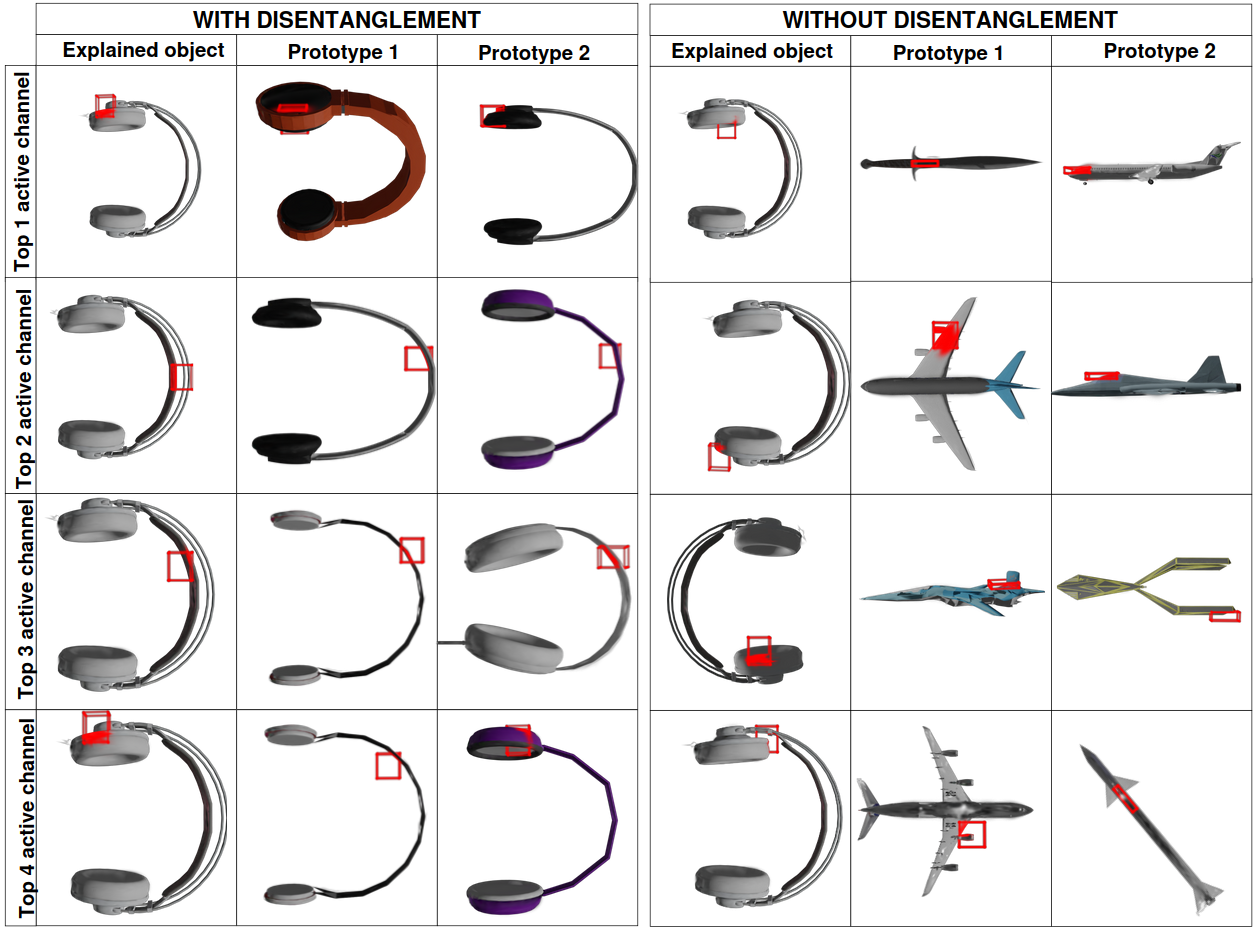}
  \caption{\textbf{Left:} Selected prototypes for the most active channels with the disentangling module applied. \textbf{Right:} Selected prototypes for the same object without applying the disentangling module.}
  \label{fig:bef-aft-didentanglement}
\end{figure}

\section{Experiments}

\label{sec:experiments}

In this section, we validate the effectiveness of XSPLAIN through a comprehensive evaluation involving both qualitative visualizations, quantitative fidelity checks and user study. We utilize a diverse set of data, including a custom version of the synthetic Toys4K dataset \citep{Stojanov2021UsingST}, which we converted to 3DGS using TRELLIS \citep{xiang2024structured} (referred to as ``Toys''), a subset of ShapeSplat, and the 3D Real Car Toolkit \citep{du20243drealcar}, to assess the method’s ability to capture semantic attributes and distinguish between fine-grained geometric features. In Table~\ref{tab:gs_pointcloud_benchmark}, we report baseline results for several supervised point cloud architectures, namely PointNet \citep{qi2017pointnetdeeplearningpoint}, PointNet++ \citep{qi2017pointnet++}, PointNeXt \citep{qian2022pointnext} and PointMLP \citep{ma2022rethinking} after incorporating Gaussian coefficients on these datasets.

\paragraph{Prototype Visualization and Interpretability}

Figure~\ref{fig:bef-aft-didentanglement} visualizes the behavior of our method on an object from the ShapeSplat dataset, comparing prototype selection with and without the use of the disentangling matrix. 
For a given target object and a specific feature channel, we identify and highlight the 3DGS voxel with the highest activation for that channel. 
Each row shows the target object alongside the corresponding fragments from prototype objects that are maximally activated for the same channel.
The left panel presents results obtained using the proposed disentangling matrix, whereas the right panel shows the same procedure without applying this matrix.
As observed, incorporating the disentangling matrix leads to a markedly stronger semantic and geometric correspondence between the target fragments and the selected prototypes, indicating that the proposed method more effectively aligns channel-specific geometric attributes across different object instances.

\begin{figure}[t]
  \centering
  \includegraphics[width=\columnwidth]{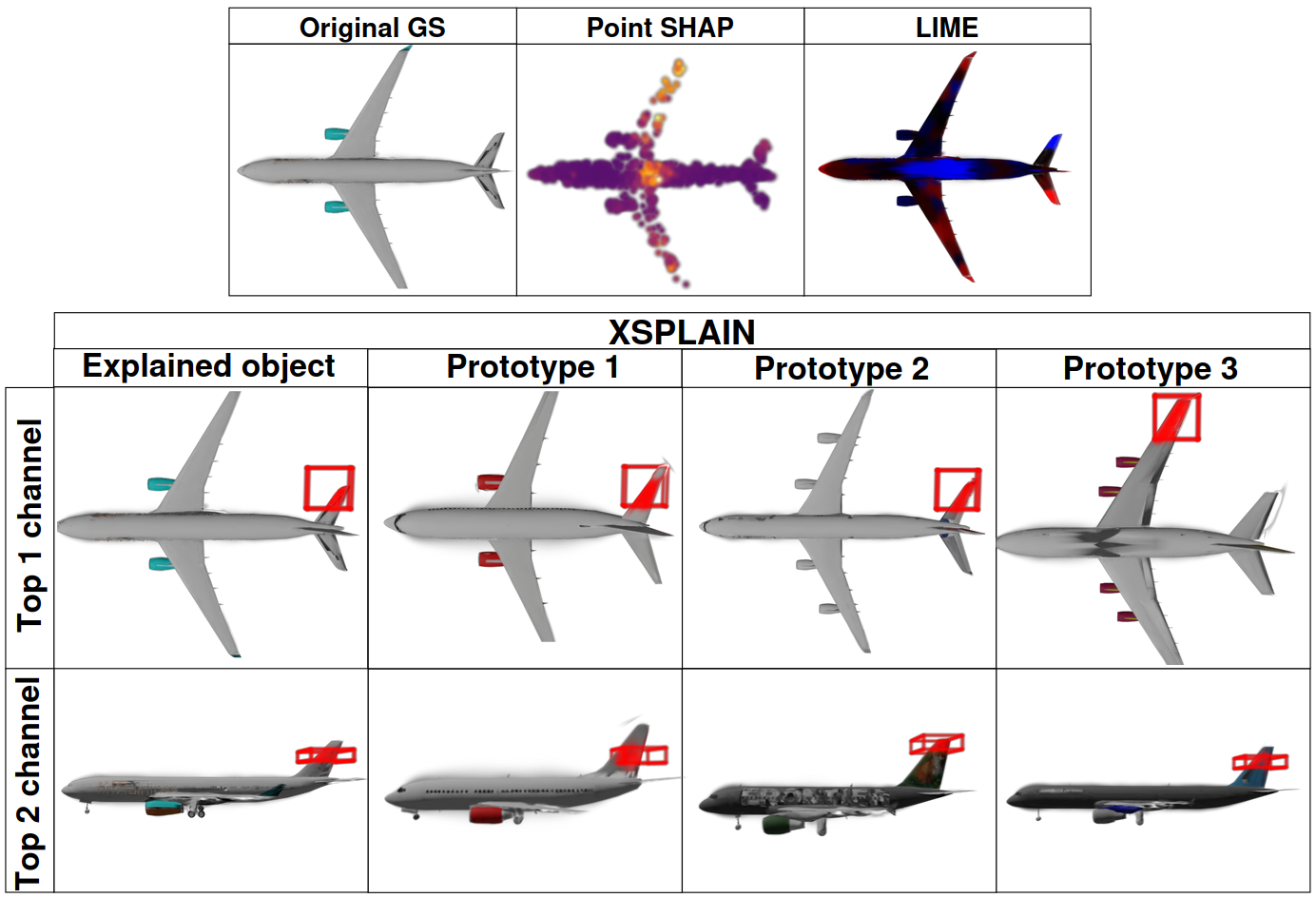}
  \caption{Comparison between PointSHAP, LIME and XSPLAIN explanations}
  \label{fig:pointshap-lime}
\end{figure}

\noindent \textbf{Comparison with Post-Hoc methods} We compare XSPLAIN against established post-hoc explanation methods adapted for point-based representations, specifically PointSHAP and LIME. As shown in Figure~\ref{fig:pointshap-lime}, baseline methods often produce scattered or noisy attributions that fail to respect the underlying object structure. In contrast, XSPLAIN leverages the voxel aggregation mechanism to produce coherent, localized explanations that align better with human intuition regarding object parts. Additional qualitative examples across different object categories and datasets are provided in the appendix \ref{app:more_examples}.
\begin{figure}[t]
  \centering
  \includegraphics[width=\columnwidth]{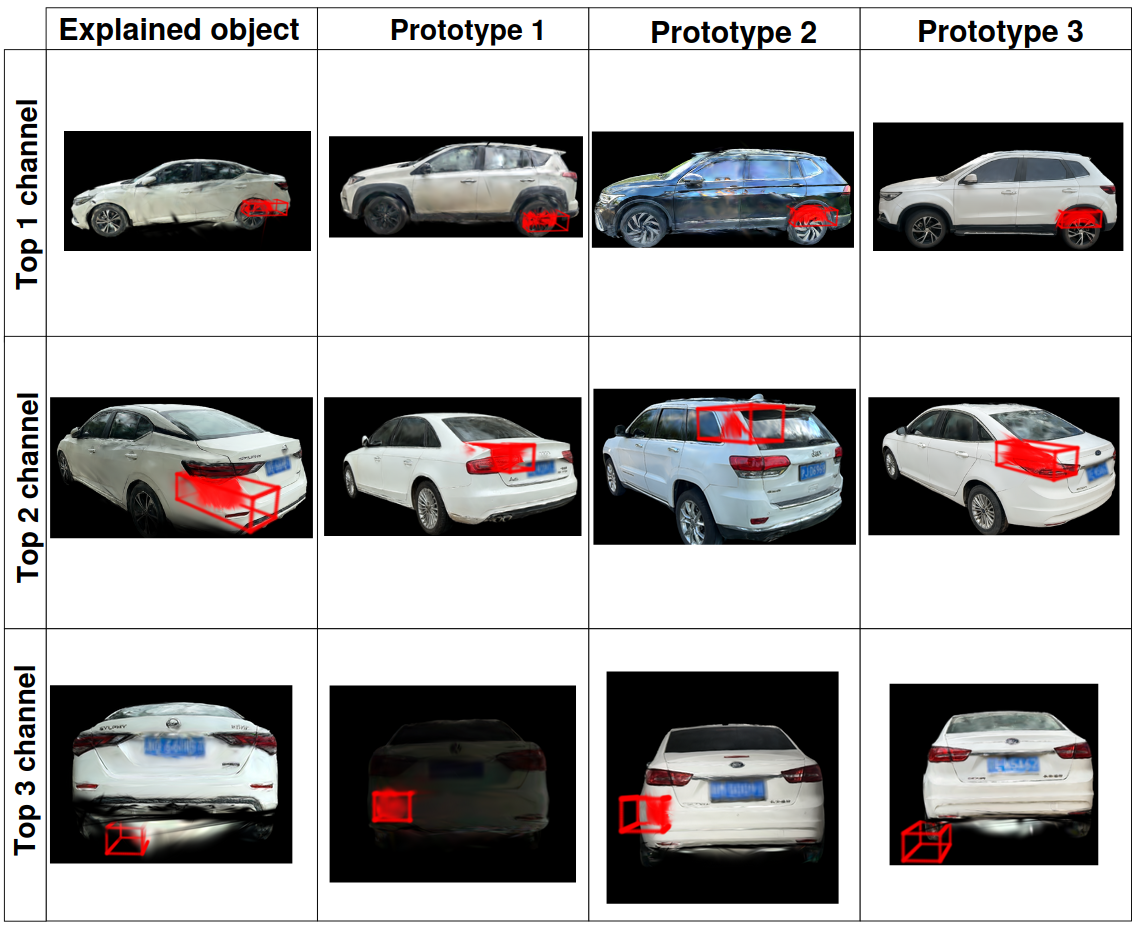}
  \caption{XSPLAIN results on samples from the 3D car dataset.}
  \label{fig:cars_explaining}
\end{figure}

\noindent \textbf{Semantic Discrimination} We further evaluate the model's capability to discern subtle geometric differences using real-world datasets.
To test the method's sensitivity to fine-grained features within a superclass, we performed experiments on the 3D Real Car Toolkit \citep{du20243drealcar}. Specifically, we analyzed the model's ability to distinguish between different vehicle types, such as SUVs and Sedans. Our visualizations in Figure ~\ref{fig:cars_explaining} show that the learned features focus on type-specific geometry (e.g., the roofline height or trunk shape), demonstrating that XSPLAIN effectively disentangles intra-class variations crucial for precise classification.

\noindent \textbf{Decoupled vs. Joint Training} To validate our 2-stage design choice, we compared \our{} against an end-to-end (one-stage) training strategy. In the 1-stage baseline, the purity loss is minimized concurrently with the classification loss using a mixed-batch approach, where part of each batch consists of prototype candidates.
Our experiments reveal that the 1-stage approach leads to a lower classification accuracy than to our decoupled method and prototype selection as shown in \autoref{tab:joint_vs_decoupled}. By freezing the backbone, we ensure that the decision boundaries remain optimal, while the orthogonal transformation $\mathbf{U}$ effectively aligns the latent space for interpretability without sacrificing model performance.

\begin{figure}[t]
  \centering
  \includegraphics[width=\columnwidth]{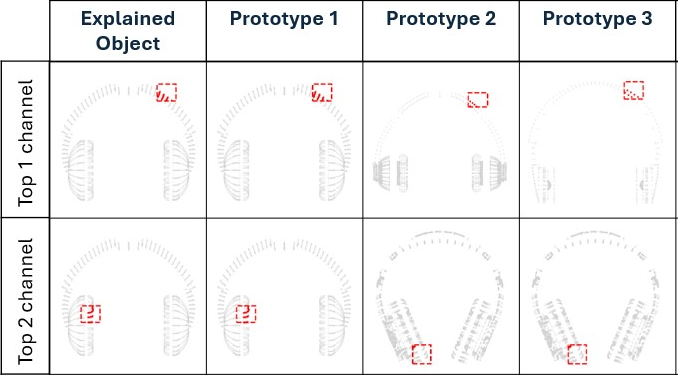}
  \caption{\textbf{Generalization to ShapeNet Core.} The model identifies distinct semantic regions in the point cloud, clearly distinguishing the headband (Row 1) and the ear cups (Row 2).}
  \label{fig:shapenet_explaining}
\end{figure}

\noindent \textbf{Direct Comparison with Point Clouds} A key advantage of \our{} is its versatility. Our method is not limited to 3D Gaussian Splatting but generalizes effectively to standard point cloud data. To validate this, we evaluated our method on the \textbf{ShapeNet Core} dataset~\cite{shapenet2015} using only geometric features ($XYZ$ coordinates and normals). As shown in Figure~\ref{fig:shapenet_explaining}, the model successfully captures semantic structures (e.g., distinguishing the headband and the ear cups) purely from geometry. 

\begin{table}[t]
    \centering
    \begin{tabular}{lccc}
    \hline
    \multirow{2}{*}{Method} & \multicolumn{2}{c}{Hyperparameters} & \multirow{2}{*}{Acc.} \\ \cline{2-3}
     & Pool ($\rho$) & $\lambda_{\text{purity}}$ \\
    \hline
    $\text{Baseline}^{\star}$ & - & - & \textbf{0.880} \\ 
    $\textbf{XSPLAIN}^{\ast}$ & - & - & \textbf{0.880} \\ 
    Joint Opt. & 30\% & 0.5 & 0.660 \\ 
    Joint Opt. & 30\% & 1.0 & 0.726 \\ 
    Joint Opt. & 50\% & 0.5 & 0.625 \\ 
    Joint Opt. & 50\% & 1.0 & 0.690 \\ \hline
    \end{tabular}
    \vspace{0.5ex}
    \caption{\textbf{Impact of training strategy on model performance.} The two-stage approach preserves classification accuracy while improving interpretability, whereas joint optimization (1-stage) fails to balance these objectives. \textbf{Note:} $\lambda_{\text{purity}}$ weights the interpretability loss, and $Pool(\rho)$ denotes the fraction of training samples used as candidate prototypes. $\text{Baseline}^{\star}$ represents Raw Backbone and $\textbf{XSPLAIN}^{\ast}$ Decoupled training.} 
    \label{tab:joint_vs_decoupled}
\end{table}

\begin{figure}[t]
  \centering
  \includegraphics[width=\columnwidth]{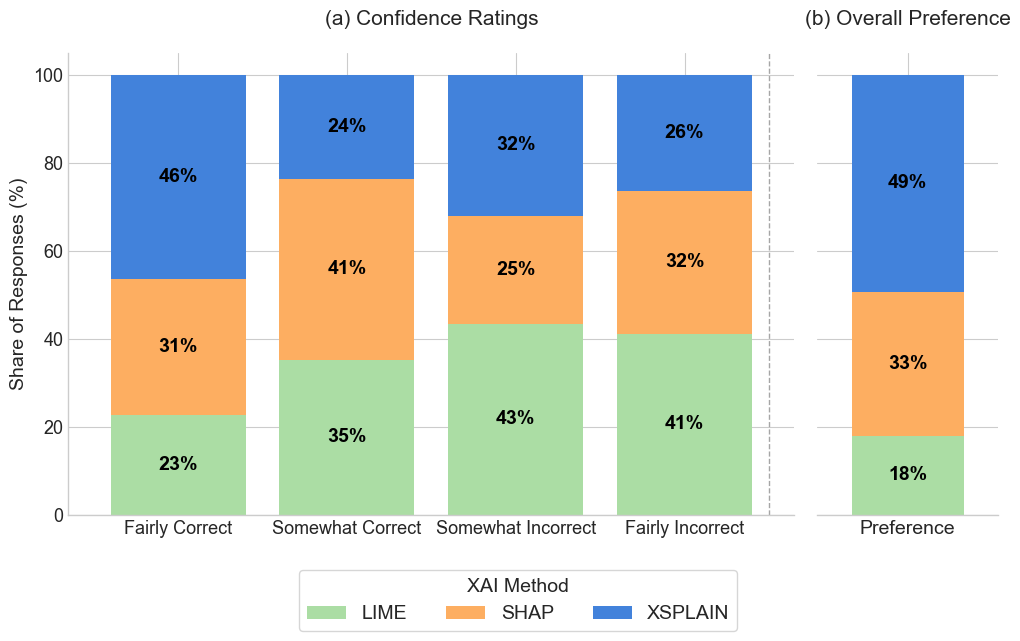}
  \caption{Results of the user study. Question (a) shows the share of each method within specific confidence ratings, highlighting that \our{} dominates the highest confidence category. Question (b) shows the direct user preference share, with \our{} selected half the time.}
  \label{fig:pool_results_plot}
\end{figure}

\noindent \textbf{Poll results analysis} The poll was performed on $N=51$ respondents with various background and experience in Machine Learning and XAI domains. LIME \citep{ribeiro2016should}
 and SHAP \citep{lundberg2017unified} were chosen as competitive methods for two reasons: (i) they provide visual heatmaps that underline regions important for the prediction, in a similar way that \our{} selects most active regions, and (ii) their prevalence in different domains make them a suitable first-choice for any explainability task. 
The questions varied between two types: (a) the perceived confidence regarding model's prediction for each method and (b) single-choice preference from proposed explanations. Fig.~\ref{fig:pool_results_plot} shows the overall qualitative results with share in each question for categories in (a) and preference for (b). The results showed a significant preference for \our{} with more than 49\% of the responses selecting it. More detailed statistical analysis can be found in~Appendix~\ref{statistical_pool_analysis}.

\noindent \textbf{Faithfulness and Ablation Study}
To objectively quantify the faithfulness of our explanations, we performed deletion tests (detailed in Appendix~\ref{app:deletion_test}). By masking the most active voxels associated with the predicted class, we observed a measurable drop in accuracy across all datasets, notably a 6.82\% degradation on Toys~\citep{Stojanov2021UsingST} and 6.13\% on MACGS~\citep{zhang2025mitigating} when removing top-5 voxels. This confirms that the regions identified by \our{} are indeed critical for the model's decision making. Finally, our ablation study (Appendix~\ref{app:hyperparameters}) highlighted the critical role of density-aware regularization ($\lambda_{den}$), which significantly boosts prototype purity by shifting focus from sparse outliers to coherent structures. We further identified a voxel grid resolution of $G=7$ as the effective balance point between spatial specificity and robustness to sparsity. For the feature dimension, while lower dimensions performed well, we selected $C=256$ to ensure sufficient latent capacity for disentangling complex semantic attributes.

\section{Conclusion}
We presented \our{}, an ante-hoc prototype based explainability framework tailored to 3D Gaussian Splat classification that provides spatially grounded and example based rationales without changing the underlying classifier decisions, hence keeping its efficacy. The method combines a voxel aggregated PointNet-style backbone with a trainable orthogonal feature rotation and an exact classifier compensation, which together preserve decision boundaries by construction, while reshaping the latent space toward channel wise purity and prototype retrieval. Across evaluated datasets, \our{} maintained the same classification accuracy as the frozen backbone, including $0.880$ on ShapeSplat, while improving interpretability as evidenced by large gains in prototype purity and coherent part level localization. In a user study with $N=51$, users selected \our{} explanations as the best across 3 methods in $48.4\%$ of cases with a statistically significant preference over adapted post-hoc baselines with $p<0.0001$, indicating higher perceived transparency under realistic inspection conditions. Faithfulness was further supported by deletion tests where removing the top activated voxels reduced accuracy by up to $6.82\%$ on Toys data (a custom 3DGS conversion of Toys4K~\cite{Stojanov2021UsingST}), and by $2.36\%$ after removing only the single most influential voxel on MACGS~\cite{zhang2025mitigating}, showing that the highlighted regions carry measurable predictive signal. These results position \our{} as a practical path to interpretable 3DGS recognition that aligns human understandable evidence with model internal mechanisms while avoiding the instability and visual noise typical of saliency-based XAI. 
{Future work} may extend the framework to richer and more diverse 3DGS corpora, incorporate appearance and view-dependent attributes when available, and developing  stronger quantitative faithfulness and robustness protocols that evaluate explanation stability under controlled perturbations of primitives and under distribution shift.  Moreover, a counterfactual~\cite{10.1145/3677119} interpretability investigation would strengthen the claims.

\paragraph{Limitation} Note that existing datasets for Gaussian Splatting classification are limited to simple objects~\cite{zhang2025mitigating}, but the fast development of GS-based models increases their importance in a high pace, allowing the method to be tested on a more varied datasets in the future.






\section{Acknowledgements}
P. Spurek was supported by the project Effective Rendering of 3D Objects Using Gaussian Splatting in an Augmented Reality Environment (FENG.02.02-IP.05-0114/23), carried out under the First Team programme of the Foundation for Polish Science and co-financed by the European Union through the European Funds for Smart Economy 2021–2027 (FENG). The work of P. Borycki was supported by the National Centre of Science (Poland) Grant No. 2025/57/N/ST6/04389. This work has been partially funded by Department of Artificial Intelligence, Wrocław University of Science and Technology. We gratefully acknowledge Poland's high-performance Infrastructure PLGrid WCSS for providing computer facilities and support within computational grant no. PLG/2025/018654. 

\nocite{langley00}

\bibliographystyle{icml2026}

\newpage
\appendix
\onecolumn
\section{Structure of the Poll}

The user study was conducted using a Google Form interface to ensure accessibility and standardization. Participants were presented with a series of questions, each featuring two distinct types of visual stimuli displayed as animated GIFs. For every query, the interface displayed three anonymized methods: Method A, Method B, and Method C.

Unbeknownst to the participants, the labels were fixed: Method C always represented our proposed \our{} framework (incorporating the original query and prototypes), while Methods A and B represented baseline approaches.

\begin{figure}[H]
    \centering
    \includegraphics[width=0.75\linewidth]{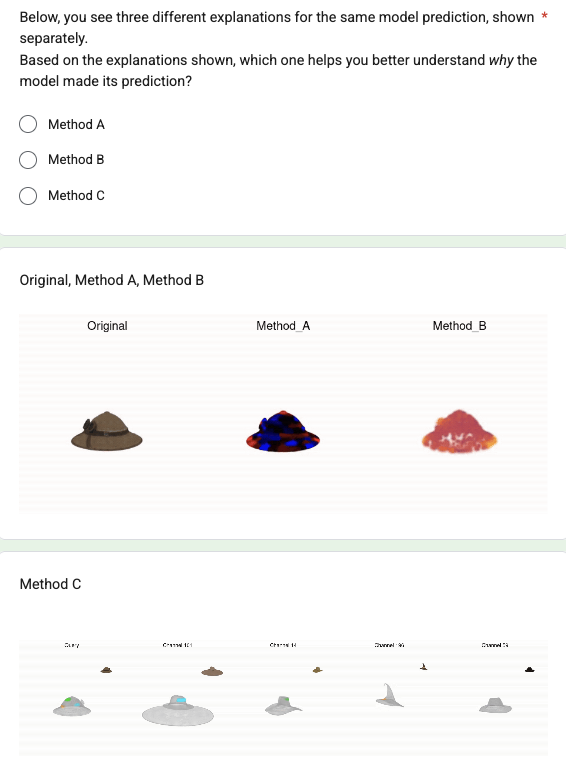}
    \caption{An exemplary view of the survey interface used to evaluate the quality of explanations across different methods.}
    \label{fig:placeholder}
\end{figure}

\begin{figure}[H]
    \centering
    \includegraphics[width=0.75\linewidth]{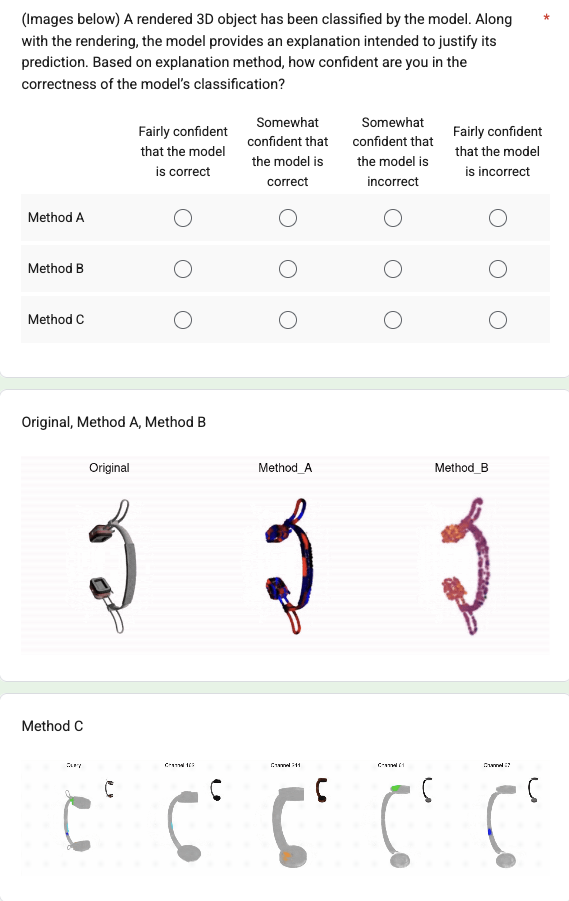}
    \caption{An exemplary view of the survey interface where participants rated how strongly the provided explanation convinced them of the classification's correctness.}
    \label{fig:placeholder}
\end{figure}

\section{Extended analysis of the responses} \label{statistical_pool_analysis}

\subsection{Data structure and inferential targets}

We collected full responses from $N = 51$ respondents. The study group consisted of 6 women and 45 men. Regarding Machine Learning expertise, the majority were intermediate $(N_{i,\mathrm{ML}}=26)$ or advanced $(N_{a,\mathrm{ML}}=14)$ practitioners, with the remainder being beginners $(N_{b,\mathrm{ML}}=9)$ or having no experience $(N_{0,\mathrm{ML}}=2)$. Experience with 3D modelling was more limited, with most participants identifying as beginners $(N_{b,\mathrm{3D}}=22)$ or having no experience $(N_{0,\mathrm{3D}}=21)$, and only 8 intermediate users. Familiarity with Explainable AI (XAI) was mixed: 23 participants had no prior experience, while the rest were intermediate $(N_{i,\mathrm{XAI}}=16)$, beginners $(N_{b,\mathrm{XAI}}=11)$, or advanced $(N_{a,\mathrm{XAI}}=1)$ users. Each respondent evaluated 3 separate items. In each item, the respondent saw 3 explanation methods, denoted Method A, Method B, and Method C. The methods behind the notation were hidden from the respondents and denoted Lime~\cite{ribeiro2016should} by A, PointSHAP~\cite{lundberg2017unified} by B, and \our{} by C. Two information were collected per item.

\begin{enumerate}
\item \textbf{Best method selection:} a single categorical choice among $\{A,B,C\}$.
\item \textbf{Confidence rating:} a four level confidence rating, mapped to integers ("Fairly confident correct" to 4,
"Somewhat confident correct" to 3, "Somewhat confident incorrect" to 2, "Fairly confident incorrect" maps to 1) that the selected explanation supports a correct understanding.
\end{enumerate}

Statistical analysis follows a deliberate sequence that mirrors these measurement properties. First, for the categorical selection outcome, the primary question is whether the population choice probabilities differ from the uniform allocation across three categories. This motivates a goodness of fit test within each item. After establishing whether non-uniformity exists in an item, the analysis proceeds to pairwise contrasts to identify which methods differ, with Holm correction to control familywise error within the set of three pairwise comparisons. Second, for confidence, the data are repeated measures because each respondent rates all 3 methods within the same item. The primary question is whether confidence differs systematically across methods within an item. This motivates a repeated measures omnibus test, specifically the Friedman test, followed by paired Wilcoxon signed rank comparisons with Holm correction only when the omnibus result is statistically significant.

Throughout, the nominal significance level is $\alpha = 0.05$. Holm correction is applied separately within each family of pairwise tests, meaning within one item for the selection comparisons and within one confidence block for the confidence comparisons.

Note that the 2 outcomes require different inferential tools as they encode different forms of replication. Within a fixed item, the $N=51$ observations are naturally treated as independent across respondents, because each respondent contributes exactly one choice for the best method. The dependence problem appears only when we stack across items, because then each respondent contributes 3 choices and those can be correlated within the same person.
For the confidence, the dependence exists already within each item, as each respondent rates all 3 methods in the same context. The 3 ratings from one respondent may not be independent, hence we include repeated measures inference, which explicitly accounts for within-respondent coupling by comparing methods of each respondent and then aggregating evidence across respondents.
To account for this we use $\chi^2$ goodness of fit for item level selections, and Friedman and Wilcoxon procedures for confidence ratings. Each method matches the sampling unit and the dependence structure of the corresponding outcome.

\subsection{Best method selection}

For each item, let $(X_A,X_B,X_C)$ denote the counts of best method selection for each of the items, across the $N=51$ respondents, naturally, $X_A+X_B+X_C = N$. The null hypothesis for each item is uniform choice probabilities
\[
H_0: p_A = p_B = p_C = \frac{1}{3}.
\]
Under $H_0$, the expected count in each category is $E_j = N/3$. The omnibus test is Pearson's $\chi^2$ goodness of fit statistic
\[
\chi^2 = \sum_{j \in \{A,B,C\}} \frac{(X_j - E_j)^2}{E_j},
\]
which is asymptotically $\chi^2$ distributed with $k-1 = 2$ degrees of freedom.

For each of the items, we obtained rejections of the hypothesis of uniformity at $\alpha=0.05$ for the two first items. The hypothesis could not be rejected for the third item.
\begin{itemize}
\item \textbf{Item 1:} $(X_A,X_B,X_C)=(9,16,26)$, $\chi^2=8.588$, $p=0.01365$. 
\item \textbf{Item 2:} $(X_A,X_B,X_C)=(8,18,25)$, $\chi^2=8.588$, $p=0.01365$. 
\item \textbf{Item 3:} $(X_A,X_B,X_C)=(11,17,23)$, $\chi^2=4.235$, $p=0.1203$.
\end{itemize}

Following a significant omnibus test, the script we performed pairwise comparisons and applied Holm correction for controlling false positives among the three pairwise contrasts for a given item. For the first item, Holm-corrected post-hoc tests found a difference between selection of methods A and C ($p=0.001176$), while comparisons involving B were not significant after correction ($p=0.107109$ for A vs B, and $p=0.088467$ B vs C). With the second item, Holm-corrected post-hoc tests found differences in selection of A versus C ($p=0.000962$) and A versus B ($p=0.046174$). The comparison B versus C was not significant after correction with $p=0.160443$.
Due to the results, we desided to perform the analysis of the joint selection in the poll.

\paragraph{Stacked selection analysis across items}

If the three items are stacked, each respondent contributes three selections, yielding $n = 3N = 153$ total selections with counts
\[
(X_A,X_B,X_C) = (28,51,74).
\]
The $\chi^2$ goodness of fit test against uniformity yields $\chi^2=20.745$ and $p = 3.128 \times 10^{-5}$, which is strong evidence against uniform choice probabilities. The stacked test treats the $n=153$ selections as independent and identically distributed. This is not strictly correct because each respondent contributes 3 observation. The strong $\chi^2$ signal and consistent directionality across items suggests robustness, but the p-value from the stacked test is approximated. The present conclusions values should be taken as strong evidence of preferential selection for Method C, with p-values understood as approximate under within respondent dependence. 
What is important, Holm correction keeps the order of magnitude for each of the compaired pairs in the stacked test, i.e., $p= 7.285604\times 10^{-8}$ after correction and $p= 2.428535\times 10^{-8}$ pre-correction for A vs C, $p= 5.321581\times 10^{-3}$ after correction and $p= 2.660791\times 10^{-3}$ pre-correction for A vs B, as well as $p= 7.476980\times 10^{-3}$ both before and after correction for B vs C, indicating that all 3 pairs differ jointly across all the items.
We decided to investigate the size ot this effect with Cram\'er's $V=\sqrt{\frac{\chi^2}{n(k-1)}}=\sqrt{\frac{20.745}{153\cdot 2}}=0.2604,$
which corresponds to a small to moderate association magnitude for a 3-category outcome.

\paragraph{Confidence intervals for selection proportions}
Let $\hat p_m = X_m/n$ for method $m \in \{A,B,C\}$ under the stacked view. The reported $95\%$ Wilson score intervals are:
\[
\hat p_A = \frac{28}{153} = 0.183,\qquad \mathrm{CI}_{0.95} = [0.130, 0.252],
\]
\[
\hat p_B = \frac{51}{153} = 0.333,\qquad \mathrm{CI}_{0.95} = [0.264, 0.411],
\]
\[
\hat p_C = \frac{74}{153} = 0.484,\qquad \mathrm{CI}_{0.95} = [0.406, 0.562].
\]
The intervals quantify uncertainty in the marginal choice rates under the stacked view \emph{under the working independence model for the stacked data}. Method C has the highest estimated selection probability and its interval lies well above $\tfrac{1}{3}$, while Method A lies below $\tfrac{1}{3}$. Note that due to within-respondent dependence mentioned earlier, they can be slightly too narrow. Note that given the number of samples the working assumption of independence is more suitable than bootstrapping.
We further analyzed  the selections of the best method in the demographic groups defined by self-reported Machine Learning experience and by self-reported XAI experience. Groups with $n<8$ were excluded from inference.\newline
\textbf{By Machine Learning experience.}
\begin{itemize}
\item Beginner, $N=9$, stacked counts $(6,6,15)$ across $n=27$ selections. Performing the omnibus test resulted in $\chi^2=6.000$, $p=0.04979$. Holm-corrected post-hoc tests indicate Method C is selected more often than A ($p=0.035983$) and more often than B ($p=0.035983$), with Wilson interval
\[
p_C=15/27=0.556,\ \mathrm{CI}_{0.95}=[0.373,\ 0.724].
\] 
Cram\'er's $V=\sqrt{6/(27\cdot 2)}= \frac{1}{3}$, showing moderate magnitude.
\item Intermediate, $N=26$, stacked counts $(16,32,30)$ across $n=78$ selections. Performing the omnibus test resulted in $\chi^2=5.846$, $p=0.05377$, not significant at $\alpha=0.05$, hence no post-hoc claims were performed.
\item Advanced, $N=14$, stacked counts $(6,10,26)$ across $n=42$ selections. Performing the omnibus test resulted in $\chi^2=16.000$, $p=0.0003355$. Holm-corrected post-hoc tests show Method C exceeds both A ($p=0.000021$) and B ($p= 0.000838$), while no significance was found in difference between A and B (pre- and post-Holm correction $p=0.266380$). Wilson interval for Method C is
\[
p_C=26/42=0.619,\ \mathrm{CI}_{0.95}=[0.468,\ 0.750].
\]
Cram\'er's $V=\sqrt{16/(42\cdot 2)}\approx 0.4364$, indicating a moderate association magnitude.
\end{itemize}

Overall, the preference for Method C is clearest in the Advanced and Beginner groups, while the Intermediate group shows a similar direction but does not reach the $\alpha=0.05$ threshold in the stacked test. Note that we can claim that explaining the model's difference might be of the most importance for those that are the most and the least experienced.
\textbf{By XAI experience.}
\begin{itemize}
\item None, $N=23$, stacked counts $(14,19,36)$ across $n=69$ selections. Omnibus $\chi^2=11.565$, $p=0.003081$. Holm-corrected post-hoc tests show Method C exceeds A ($p=0.000293$) and exceeds B ($p=0.006238$), with A versus B showing $p=0.318360$. Wilson interval for Method C is
\[
p_C=36/69=0.522,\ \mathrm{CI}_{0.95}=[0.406,\ 0.635].
\]
\item Beginner, $N=11$, stacked counts $(7,13,13)$ across $n=33$ selections. Omnibus $\chi^2=2.182$, $p=0.3359$, hence there was no evidence against uniformity.
\item Intermediate, $N=16$, stacked counts $(7,18,23)$ across $n=48$ selections. Omnibus $\chi^2=8.375$, $p=0.01518$. Holm-corrected post-hoc tests show Method A is selected less often than B ($p=0.021045$) and less often than C ($p=0.001280$), while B versus C is not significant ($p=0.302236$). Wilson intervals include
\[
p_A=7/48=0.146,\ \mathrm{CI}_{0.95}=[0.072,\ 0.272].
\]
\end{itemize}

\subsection{Confidence ratings}

Confidence ratings were assumed as numbers in Likert scale 1-4, hence are ordinal and not normally distributed in most blocks according to Shapiro results. Therefore the primary omnibus test is Friedman repeated measures, which tests equality of the marginal distributions across methods within each block.

Effect size is Kendall's $W=\chi^2/(n(k-1))$ with $k=3$, with balues near $0$ indicating weak consistency of method ranking across respondents, values closer to $1$ indicate strong agreement.

\begin{itemize}
\item Confidence on 1st item, $n=51$, Friedman $\chi^2=2.872$, $p=0.2379$, $W=0.028$. This suggests no significant evidence of differences in the subject reported confidence across methods.
\item Confidence on 2nd item, $n=51$, Friedman $\chi^2=12.896$, $p=0.001584$, $W=0.126$. Post-hoc Wilcoxon tests with Holm correction indicate a significant difference between Method A and Method C, with median difference A minus C equal to $-1$. This indicates higher confidence under Method C than under Method A by about one Likert level in median. Other pairwise comparisons were not significant after correction.
\item Confidence on 3rd item, $n=51$, Friedman $\chi^2=2.747$, $p=0.2532$, $W=0.027$. Showed no no significant evidence of differences in the subject reported confidence across methods.
\end{itemize}

Additionally, a per respondent mean confidence across blocks was computed for each method and tested with Friedman. With $n=51$, Friedman $\chi^2=3.841$, $p=0.1465$, $W=0.038$, indicating no detectable overall shift in confidence when averaging across the three items.

For paired confidence ratings, effect sizes were quantified using the Hodges-Lehmann estimator of the paired shift. For each respondent and item, paired differences were computed as
\begin{equation*}
d_i = \text{Confidence}_{i,A} - \text{Confidence}_{i,C}, \quad i=1,\dots,51.    
\end{equation*}
The HL estimator is defined as the median of the paired differences $\widehat{\Delta}_{HL} = \mathrm{median}(d_1,\dots,d_{51})$, and estimates the shift in confidence between Method~A and Method~C on the Likert scale. Uncertainty was quantified using nonparametric bootstrap confidence intervals obtained by resampling respondents with replacement and recomputing the HL estimator. For the second item, the HL estimator was $\widehat{\Delta}_{HL}=-1$ with a $95\%$ confidence interval of $[-1,\,0]$. This indicates that the typical respondent reported approximately one Likert level higher confidence under Method~C than under Method~A. The confidence interval excludes positive values and is consistent with the significant Friedman omnibus test and the Holm-corrected Wilcoxon signed-rank test for this item. While for the other two items we calculated $\widehat{\Delta}_{HL}=0$ indicating no significant differences in the confidance, as suggested by the non-significant Friedman test result. The 95\% confidence interval was equal to $[-1,\,1]$ in the first item, and $[-1,\,0]$ in the third.

\textbf{Subgroup confidence results.}
Subgroup analyses mostly mirror the overall pattern. For Machine Learning and Deep Learning Intermediate group, 2nd item shows Friedman significance with $p=0.002271$ and $W=0.234$, and the post-hoc A versus C comparison remains significant after Holm correction with median difference $-1$. Other blocks and the aggregated analysis are not significant. Whereas for Machine Learning Advanced and Beginner groups, no confidence block shows a Holm-corrected significant pairwise difference. 

For groups divided by XAI familiarity, None, Intermediate, and Beginner groups showed no Holm-corrected pairwise differences in confidence scale, although XAI Intermediate has an omnibus Friedman $p=0.02538$ in the second item that does not translate to Holm-corrected pairwise significance. 

\section{Classifier Compensation Proof}
\label{app:classifier_compensation}

We show that the classifier compensation preserves the original model predictions exactly. Let $\tilde{\mathbf{z}} = \mathbf{U}\mathbf{z}$ denote the transformed global feature,
where $\mathbf{U}$ is the orthogonal disentanglement matrix and $\mathbf{z}$ is the original global feature vector.

After compensation, the classifier weights become $\mathbf{W}'_{\text{cls}} = \mathbf{W}_{\text{cls}} \mathbf{U}^{T}$. The prediction on the transformed features is:
\begin{align}
\mathbf{W}'_{\text{cls}} \tilde{\mathbf{z}} &= (\mathbf{W}_{\text{cls}} \mathbf{U}^{T}) (\mathbf{U} \mathbf{z}) \\
&= \mathbf{W}_{\text{cls}} (\mathbf{U}^T \mathbf{U}) \mathbf{z} \\
&= \mathbf{W}_{\text{cls}} \mathbf{I} \mathbf{z} \\
&= \mathbf{W}_{\text{cls}} \mathbf{z},
\end{align}
where the third equality follows from the orthogonality of $\mathbf{U}$ ($\mathbf{U}^T \mathbf{U} = \mathbf{I}$).

Thus, the logits and consequently the predicted class remain identical before and after applying the disentanglement transformation, ensuring that the interpretability module does
not affect the model's decisions.

\section{Additional Details on Point Cloud Generalization}
\label{app:pointclouds}

Expanding on the generalization capabilities discussed in the main text, here we provide further quantitative and qualitative details regarding the experiments on standard point clouds. While our primary evaluation focuses on the rich 3D Gaussian Splatting representation, verifying performance on the \textbf{ShapeNet Core} dataset~\cite{shapenet2015} demonstrates that \our{} is not limited to 3D Gaussian Splatting objects.

\textbf{Experimental Setup.}
To isolate geometric understanding from rendering attributes, we trained the model on raw point cloud data consisting of $N$ points. The input features were limited to spatial coordinates and surface normals ($N \times 6$ features: $x, y, z, n_x, n_y, n_z$). \textbf{We utilized the same default hyperparameter configuration as in our main experiments: density regularization $\lambda=3.5$, voxel grid size $G=7$, and latent feature dimension $C=256$.}

\textbf{Quantitative Results.}
The model demonstrated high stability and effectiveness in this setting. The PointNet backbone achieved a \textbf{classification accuracy of 94.4\%}. The \textbf{purity metric} increased from \textbf{13.0\%} to \textbf{22.5\%} during the training of the orthogonal matrix.

\begin{figure}[H]
    \centering
    \includegraphics[width=0.6\linewidth]{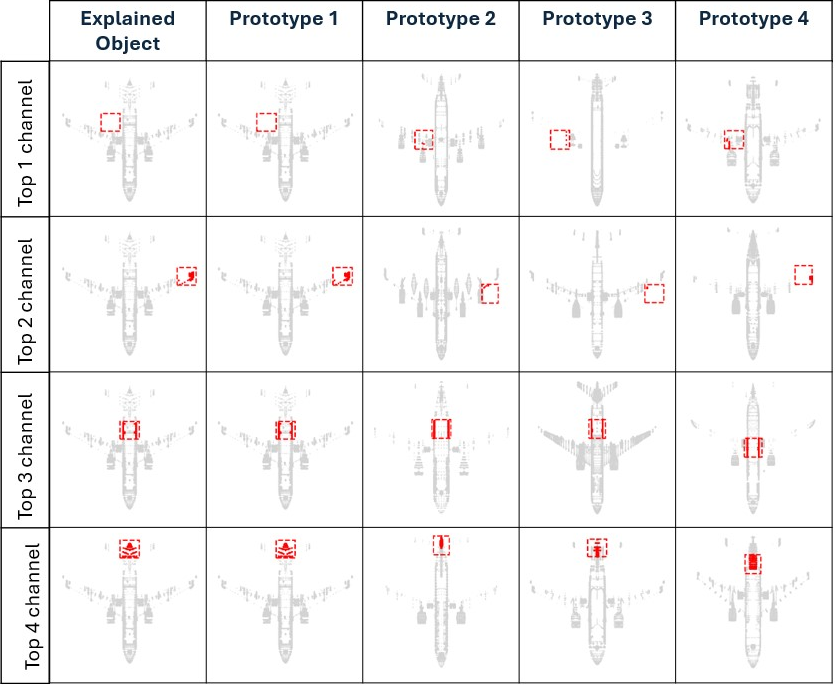}
    \caption{\textbf{XSPLAIN explanation} illustrating four prototypes corresponding to the four most active channels for an object from the \textit{Airplane} class in the ShapeNet Core dataset.}
    \label{fig:airplane_shapenet}
\end{figure}

\begin{figure}[H]
    \centering
    \includegraphics[width=0.6\linewidth]{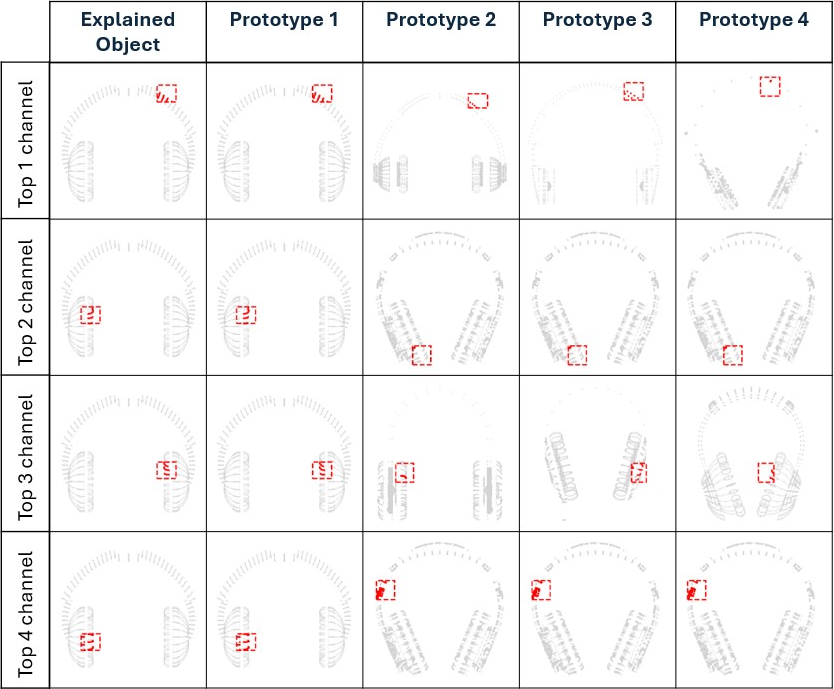}
    \caption{\textbf{XSPLAIN explanation} illustrating four prototypes corresponding to the four most active channels for an object from the \textit{Earphones} class in the ShapeNet Core dataset.}
    \label{fig:headphone_shapenet}
\end{figure}

\begin{figure}[H]
    \centering
    \includegraphics[width=0.6\linewidth]{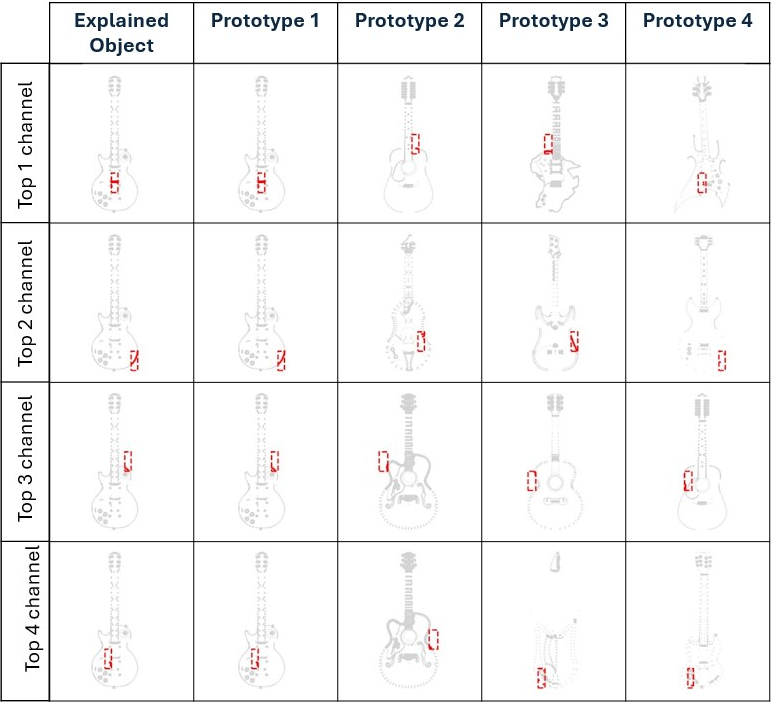}
    \caption{\textbf{XSPLAIN explanation} illustrating four prototypes corresponding to the four most active channels for an object from the \textit{Guitar} class in the ShapeNet Core dataset.}
    \label{fig:guitar_shapenet}
\end{figure}

\begin{figure}[H]
    \centering
    \includegraphics[width=0.6\linewidth]{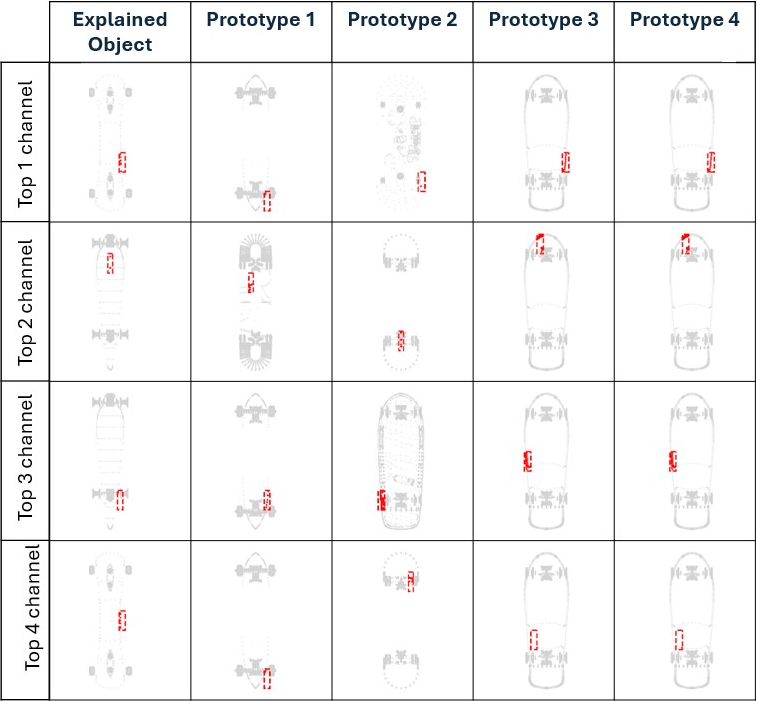}
    \caption{\textbf{XSPLAIN explanation} illustrating four prototypes corresponding to the four most active channels for an object from the \textit{Skateboard} class in the ShapeNet Core dataset.}
    \label{fig:skateboard_shapenet}
\end{figure}

\textbf{Qualitative Results.}
Visualizations of the learned prototypes confirm that the method successfully captures semantic parts based purely on geometry. As shown in Figures~\ref{fig:airplane_shapenet}, \ref{fig:headphone_shapenet}, \ref{fig:guitar_shapenet} and~\ref{fig:skateboard_shapenet}, the active channels clearly distinguish between structural components, such as separating airplane wings or the ear cups of headphones, without relying on color or opacity cues.

\newpage

\section{Additonal Explaining Examples}
\label{app:more_examples}
In this section, we present extended qualitative comparisons between \our{} and post-hoc baseline methods (PointSHAP and LIME). We visualize results across diverse object categories selected from three distinct datasets used in our main evaluation: \textbf{ShapeSplat}~\citep{ma2024shapesplat}, the \textbf{Toys} dataset (a custom 3DGS conversion of Toys4K~\citep{Stojanov2021UsingST}), and the \textbf{3D Real Car Toolkit}~\citep{du20243drealcar}.

\begin{figure}[H]
    \centering
    \includegraphics[width=0.75\linewidth]{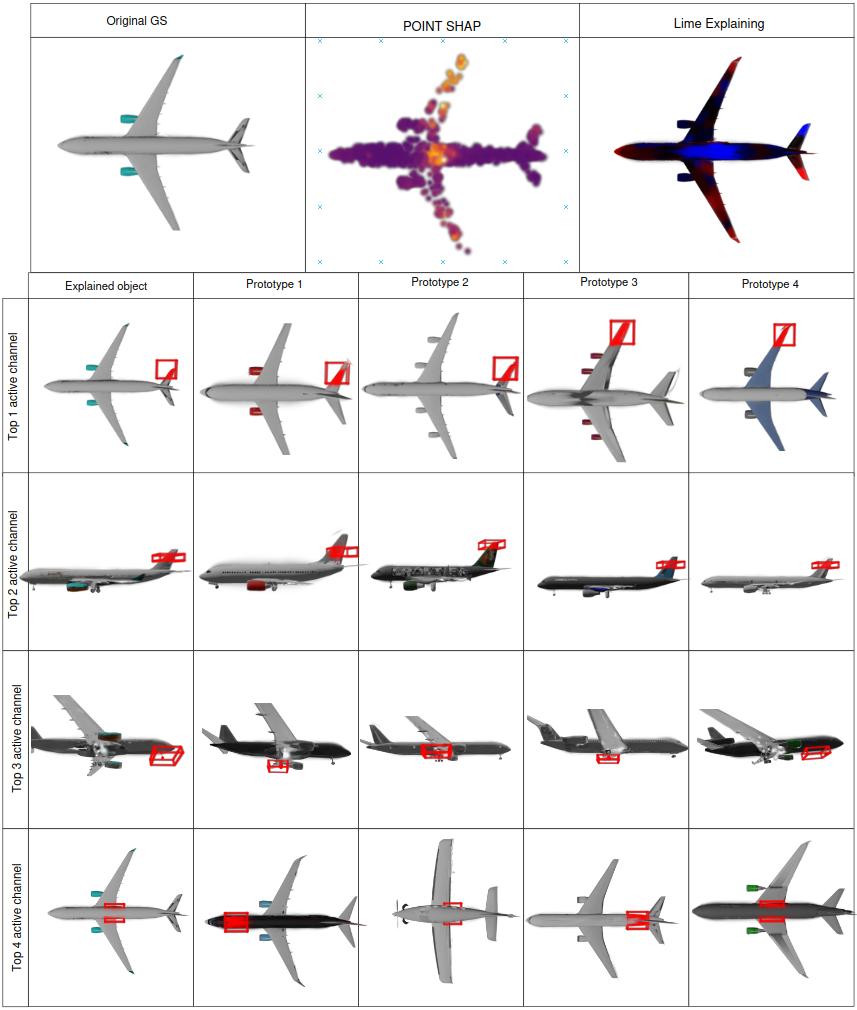}
    \caption{XSPLAIN explanation using four prototypes corresponding to the four most active channels, compared with PointSHAP and LIME for an object from airplane class from the ShapeSplat dataset.}
    \label{fig:airplane_explaining}
\end{figure}

\begin{figure}[H]
    \centering
    \includegraphics[width=0.75\linewidth]{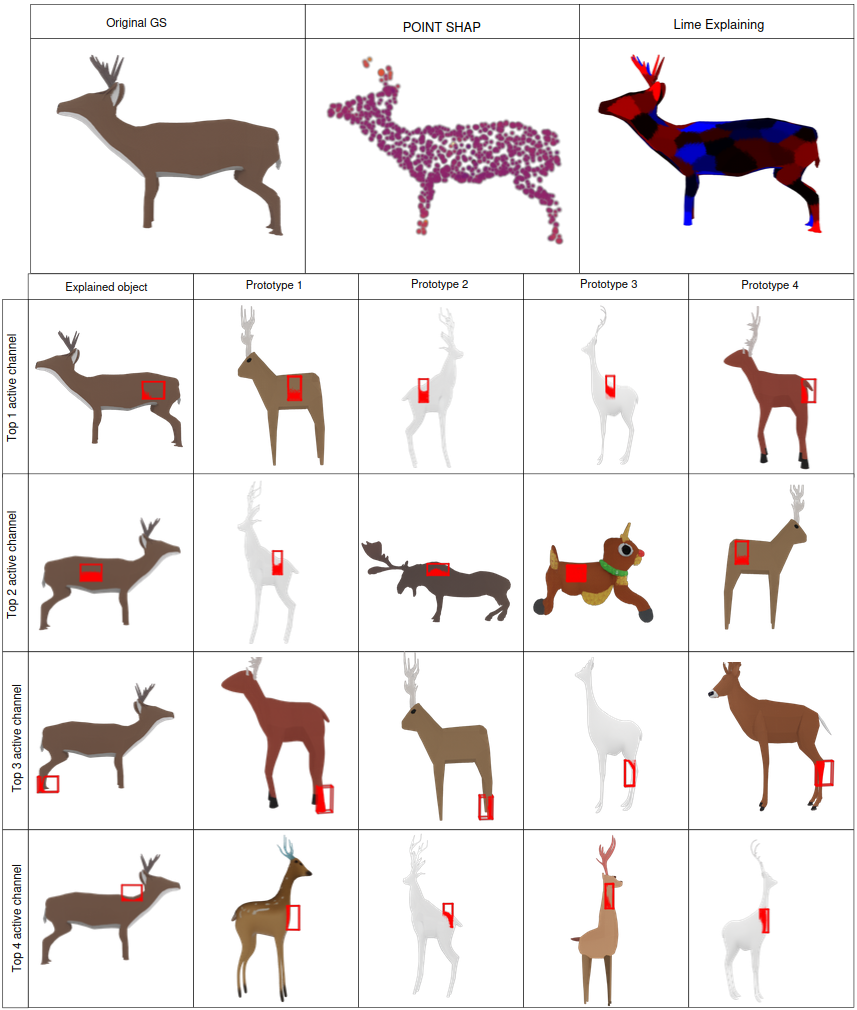}
    \caption{XSPLAIN explanation using four prototypes corresponding to the four most active channels, compared with PointSHAP and LIME for an object from deer moose class from the Toys dataset.}
    \label{fig:deer_explaining}
\end{figure}

\begin{figure}[H]
    \centering
    \includegraphics[width=0.75\linewidth]{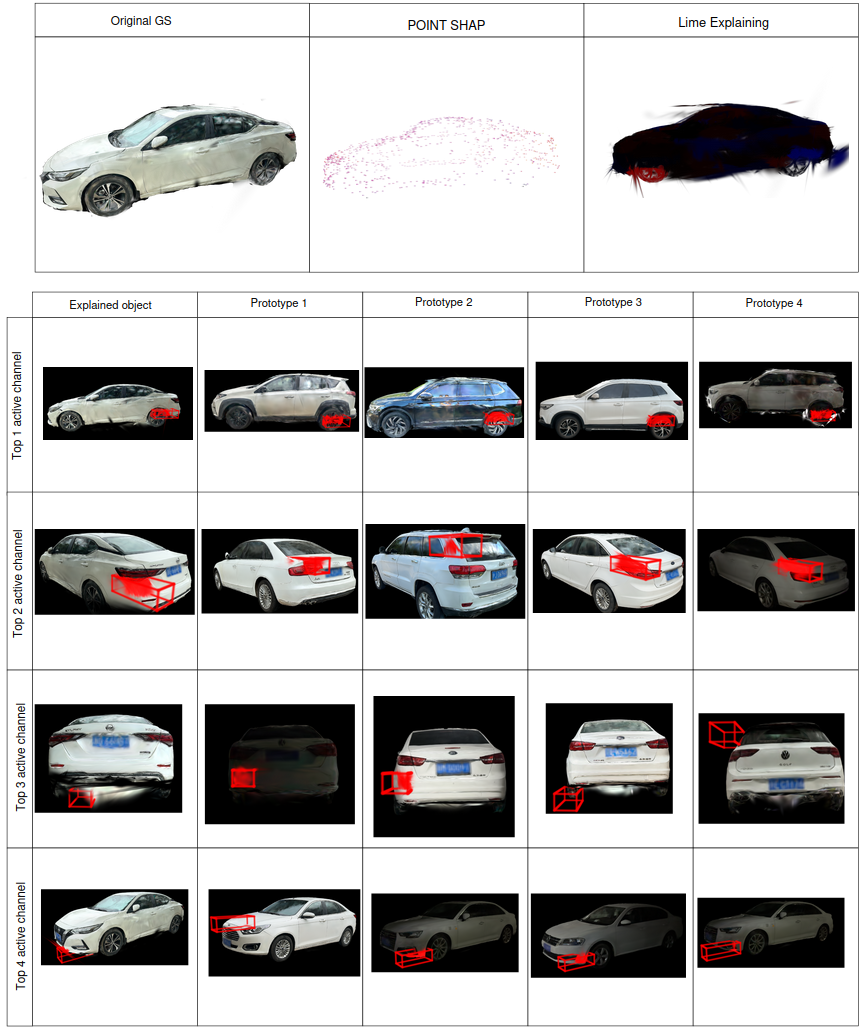}
    \caption{XSPLAIN explanation using four prototypes corresponding to the four most active channels, compared with PointSHAP and LIME for an object from the sedan class from the 3d\_cars dataset.}
    \label{fig:sedan_explaining}
\end{figure}

\begin{figure}[H]
    \centering
    \includegraphics[width=0.75\linewidth]{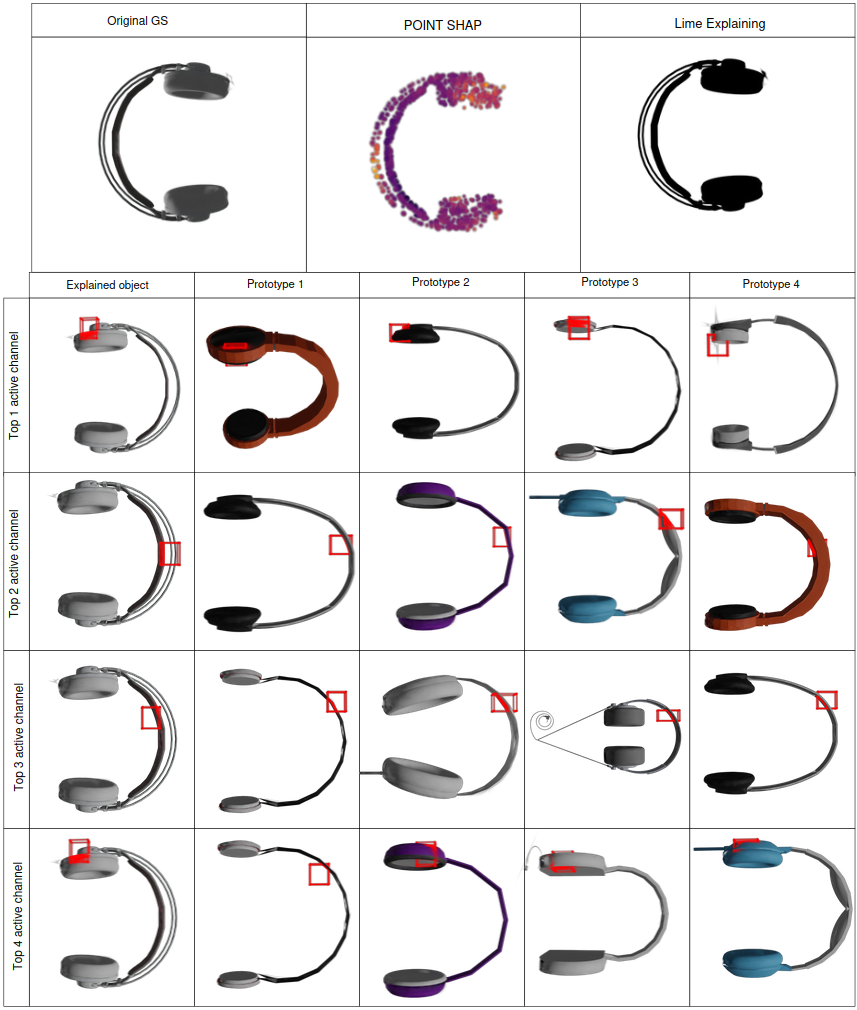}
    \caption{XSPLAIN explanation using four prototypes corresponding to the four most active channels, compared with PointSHAP and LIME for an object from earphone class from the ShapeSplat dataset.}
    \label{fig:earphone_explaining}
\end{figure}

\begin{figure}[H]
    \centering
    \includegraphics[width=0.75\linewidth]{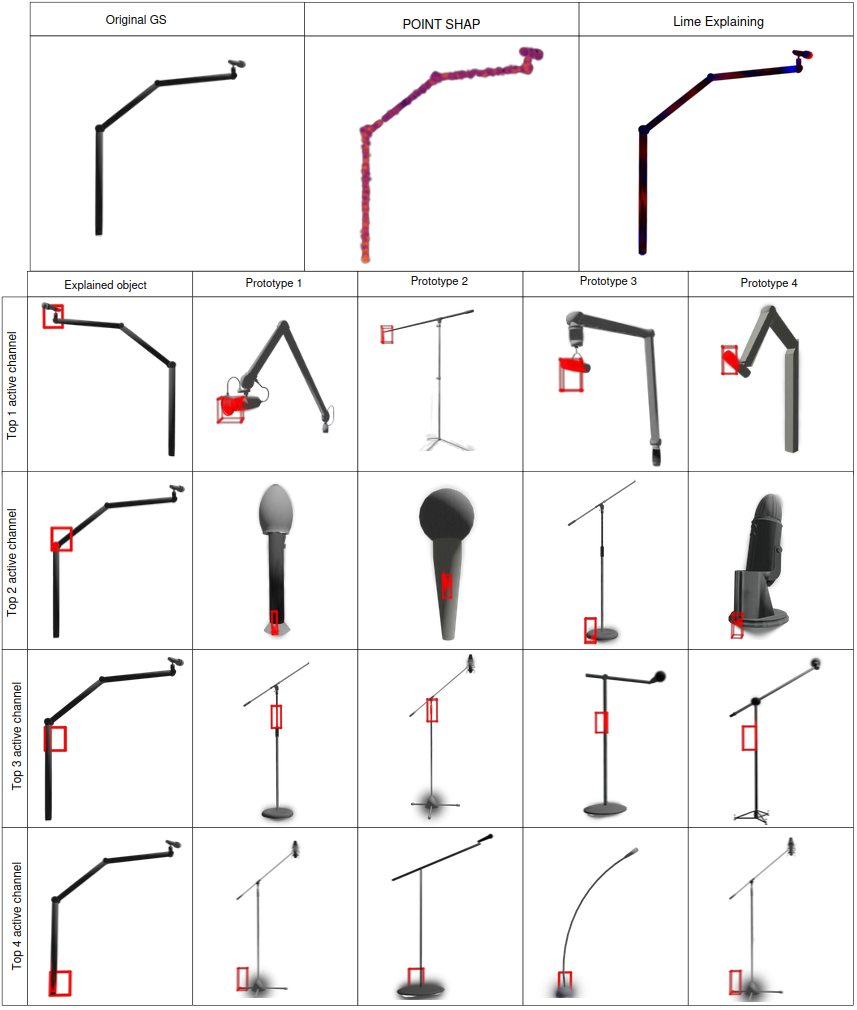}
    \caption{XSPLAIN explanation using four prototypes corresponding to the four most active channels, compared with PointSHAP and LIME for an object from microphone class from the ShapeSplat dataset.}
    \label{fig:microphone_explaining}
\end{figure}

\begin{figure}[H]
    \centering
    \includegraphics[width=0.75\linewidth]{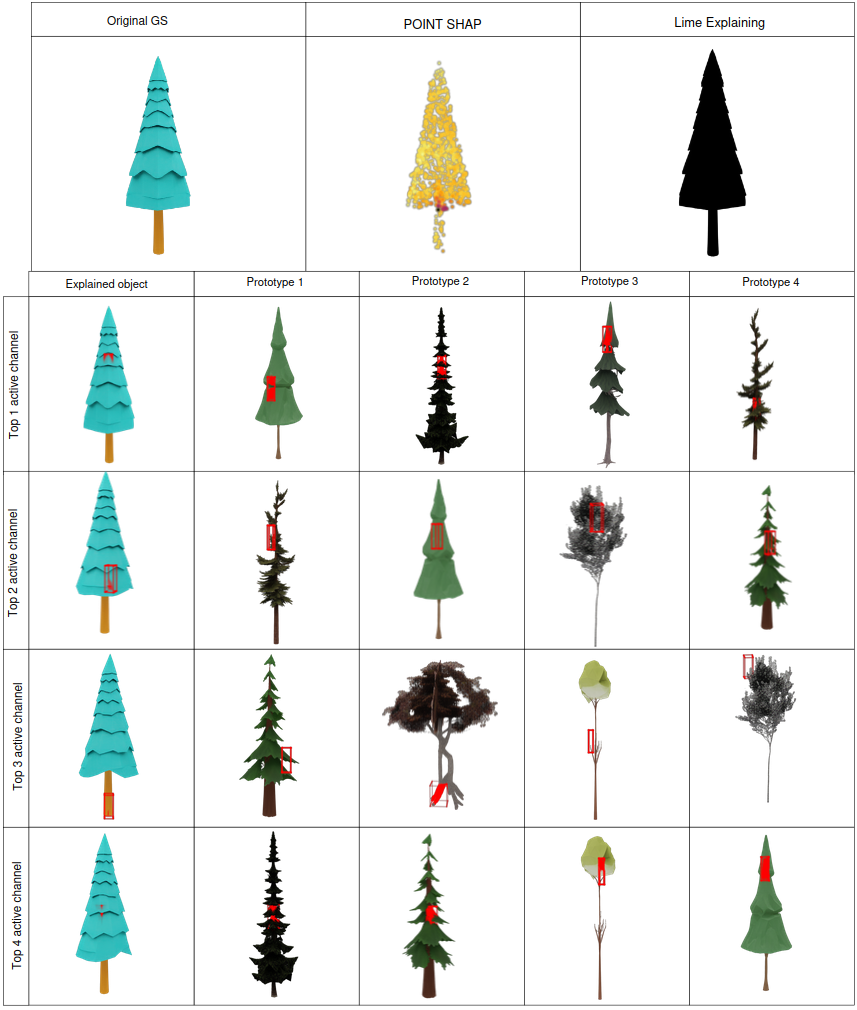}
    \caption{XSPLAIN explanation using four prototypes corresponding to the four most active channels, compared with PointSHAP and LIME for an object from tree class from the Toys dataset.}
    \label{fig:tree_explaining}
\end{figure}

\section{Deletion Test}\label{app:deletion_test}

To evaluate the importance of voxels selected for channel prototypes, we evaluated the accurracy of samples with the top k voxels, corresponding to the top k activated channels, completely removed from the input. Table \ref{tab:perturbed_comparison} shows that indeed removing important voxels influences the final accuracy. 

\begin{table}[H]
    \centering
    \begin{tabular}{lrr}
    \toprule
    Dataset & Top k & Perturbation degradation \\
    \midrule
    MACGS & 1 & 2.36\% \\
    MACGS & 2 & 5.66\% \\
    MACGS & 3 & 5.66\% \\
    MACGS & 4 & 5.19\% \\
    MACGS & 5 & 6.13\% \\
    \hline
    Shapesplat & 1 & 1.46\% \\
    Shapesplat & 2 & 2.19\% \\
    Shapesplat & 3 & 0.73\% \\
    Shapesplat & 4 & 1.46\% \\
    Shapesplat & 5 & 1.46\% \\
    \hline
    Toys & 1 & 4.55\% \\
    Toys & 2 & 4.55\% \\
    Toys & 3 & 5.68\% \\
    Toys & 4 & 4.55\% \\
    Toys & 5 & 6.82\% \\
    \bottomrule
    \end{tabular}
    \caption{Degradation in accuracy (as relative percentage change) after perturbing top k most active voxels in each sample, for each dataset.}
    \label{tab:perturbed_comparison}
\end{table}

\section{Hyperparameters}\label{app:hyperparameters}

We performed a comprehensive ablation study on the \textbf{ShapeSplat} dataset to analyze the impact of key hyperparameters on both classification performance and explanation quality. The results highlight the trade-off between pure accuracy and the interpretability metrics (Purity Gain and Voxel Density).

\textbf{Analysis}
\begin{itemize}
\item Regularization ($\lambda_{\text{den}}$): Without density regularization ($\lambda=0$), the model achieves higher accuracy but focuses on sparse outliers (Density=32), leading to poor interpretability. Increasing $\lambda$ forces the model to focus on dense, geometric regions, significantly boosting Purity Gain.
\item Grid Size ($G$): A resolution of $G=7$ offers the best balance. Coarser grids ($G=3$) lack spatial specificity, while finer grids ($G=15$) result in sparse voxels susceptible to noise.
\item Feature Dimension ($C$): While $C=64$ yields maximum accuracy, we utilize $C=256$ as the default to ensure sufficient capacity to disentangle complex semantic attributes during the interpretation stage.
\end{itemize}

\begin{table}[H]
    \centering
    \small
    \setlength{\tabcolsep}{3.5pt}
    
    \begin{tabular}{l|c|ccc}
    \toprule
    \textbf{Param.} & \textbf{Val.} & \textbf{Acc.} & \textbf{Pur. Gain} & \textbf{Dens.} \\
    \midrule
    
    \multirow{5}{*}{\makecell[l]{Reg.\\Strength\\($\lambda_{\text{den}}$)}} 
      & 0.0 & 86.3 & +37.1\% & 32 \\
      & 1.0 & 84.5 & \textbf{+86.3\%} & 78 \\
      & 2.0 & 78.5 & +67.3\% & 75 \\
      & 3.5$^\dagger$ & 82.1 & +61.1\% & 78 \\
      & 5.0 & 83.9 & +66.0\% & 79 \\
    \midrule
    
    \multirow{5}{*}{\makecell[l]{Grid\\Res.\\($G$)}} 
      & 3 & 83.3 & +55.3\% & 468 \\
      & 5 & 77.3 & +62.1\% & 158 \\
      & 7$^\dagger$ & 82.1 & +61.1\% & 78 \\
      & 10 & 81.5 & +54.9\% & 36 \\
      & 15 & 78.5 & +51.1\% & 23 \\
    \midrule
    
    \multirow{4}{*}{\makecell[l]{Feat.\\Dim.\\($C$)}} 
      & 16 & 85.7 & +36.0\% & 74 \\
      & 64 & \textbf{88.0} & +64.1\% & 74 \\
      & 256$^\dagger$ & 82.1 & +61.1\% & 78 \\
      & 1024 & 66.6 & +46.3\% & 72 \\
    \bottomrule
    \end{tabular}
    
    \caption{\textbf{Ablation Study.} \textbf{Acc.}: Classification Accuracy (\%), \textbf{Pur. Gain}: Relative improvement in prototype purity after Stage 2 optimization, \textbf{Dens.}: Mean point count in activated voxels. ($^\dagger$) indicates the default configuration.}
    \label{tab:ablation_gain}
\end{table}



\end{document}